%% file: main.tex
\documentclass[10pt,twocolumn,letterpaper]{article}

\usepackage{cvpr}              

\usepackage{algorithm}
\usepackage{algorithmic}
\usepackage{amsmath}  
\usepackage{amssymb}   
\usepackage{amsfonts}  
\usepackage{multirow}
\usepackage{booktabs, multirow} 
\usepackage{tabularx}
\usepackage{booktabs}
\usepackage{makecell}
\usepackage{siunitx}
\usepackage{pgfplots}
\usepackage[table]{xcolor}
\newcommand{\minisection}[1]{\vspace{0.005in} \noindent {\bf #1}}

\input{preamble}
\definecolor{cvprblue}{rgb}{0.21,0.49,0.74}
\usepackage[pagebackref,breaklinks,colorlinks,allcolors=cvprblue]{hyperref}

\def\confName{CVPR}
\def\confYear{2026}

\title{Language as an Anchor: Preserving Relative Visual Geometry \\ for Domain Incremental Learning}

\vspace{-3mm}

\author{
Shuyi Geng$^{1,2}$ \quad
Tao Zhou$^{3}$ \quad
Yi Zhou$^{1,2}$\thanks{Corresponding author.} \\
$^{1}$School of Computer Science and Engineering, Southeast University, China \\
$^{2}$Key Laboratory of New Generation Artificial Intelligence Technology and \\
Its Interdisciplinary Applications, Ministry of Education, China \\
$^{3}$Nanjing University of Science and Technology, China \\
{\tt\small shuyigeng@seu.edu.cn, yizhou.szcn@gmail.com}
}

\vspace{-1mm}

\begin{document}
\maketitle
\input{sec/0_abstract}    
\input{sec/1_intro}
\input{sec/2_formatting}
\input{sec/3_Preliminary_and_Method}
\input{sec/4_Experiments}

\input{sec/5_Conclusion}
{
    \small
    \bibliographystyle{ieeenat_fullname}
    \bibliography{main}
}

\input{sec/X_suppl}

\end{document}

%% file: sec/0_abstract.tex
\begin{abstract}
A key challenge in Domain Incremental Learning (DIL) is to continually learn under shifting distributions while preserving knowledge from previous domains.
Existing methods face a fundamental dilemma. On one hand, projecting all domains into a single unified visual space leads to inter-domain interference and semantic distortion, as large shifts may vary with not only visual appearance but also underlying semantics. On the other hand, isolating domain-specific parameters causes knowledge fragmentation, creating ``knowledge islands'' that hamper knowledge reuse and exacerbate forgetting.
To address this issue, we propose \textbf{LAVA} (\textbf{L}anguage-\textbf{A}nchored \textbf{V}isual \textbf{A}lignment), a novel DIL framework that replaces direct feature alignment with relative alignment driven by a text-based reference anchor. LAVA guides the visual representations of each incoming domain to preserve a consistent relative geometry, which is defined by mirroring the pairwise semantic similarities between the class names. This anchored geometric structure acts as a bridge across domains, enabling the retrieval of class-aware prior knowledge and facilitating robust feature aggregation. 
Extensive experiments on standard DIL benchmarks demonstrate that LAVA achieves significant performance improvements over state-of-the-arts.
Code is available at \url{https://github.com/ShuyiGeng/LAVA}.
\vspace{-3mm}
\end{abstract}

%% file: sec/1_intro.tex
\section{Introduction}

\begin{figure}[t]
  \centering
  \includegraphics[width=0.45\textwidth]{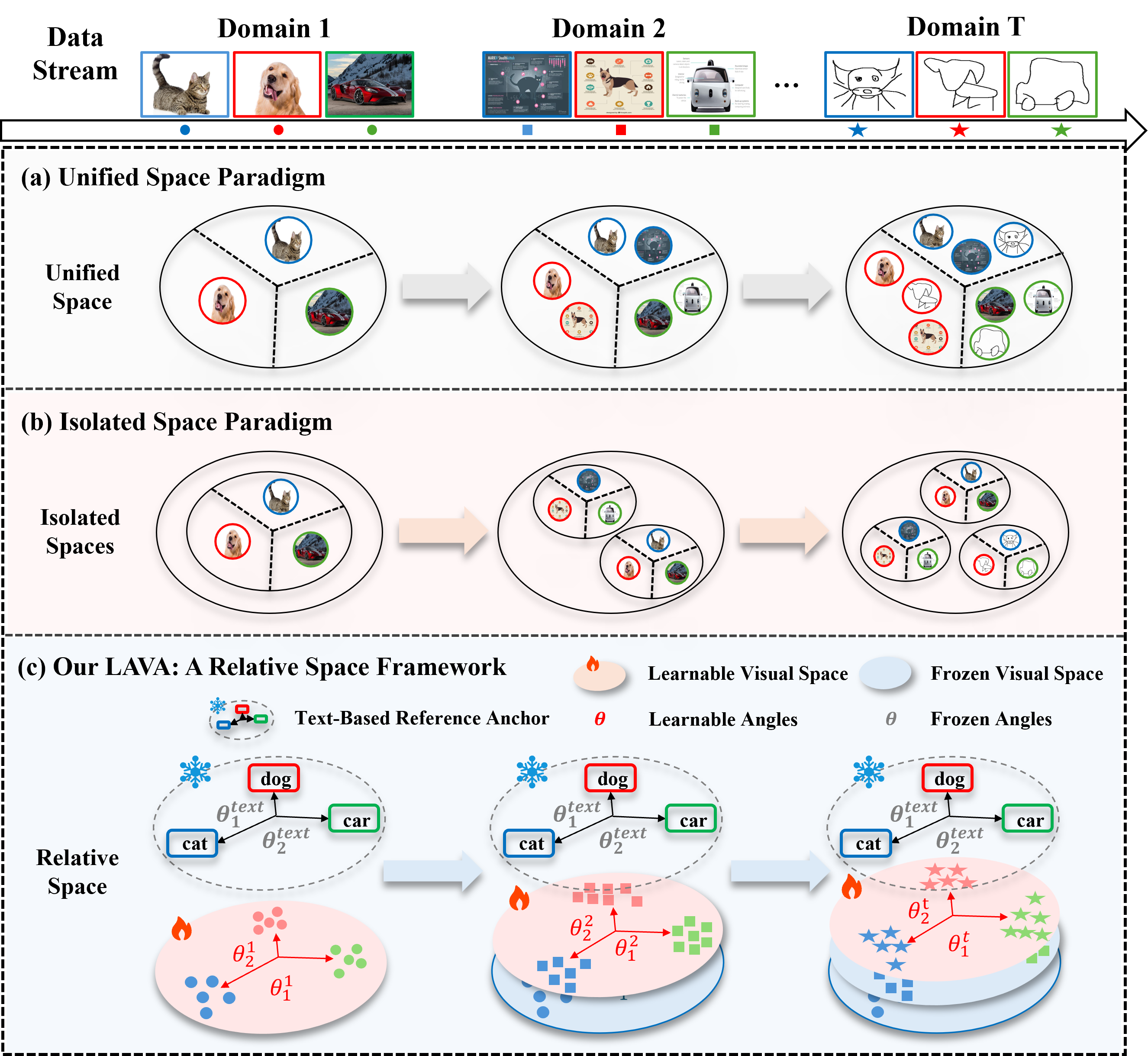}
  \caption{
    (a) Unified Space Paradigm: Projecting features from all domains into a shared visual space risks interference and semantic distortion.
    (b) Isolated Space Paradigm: Learning domain-specific subspaces via prompts or adapters leads to fragmented knowledge.
    \textbf{(c) Our LAVA: A Relative Space Framework.} LAVA aligns each domain’s relative visual geometry (e.g., the learnable visual angles $\theta^t$) with a stable semantic structure derived from a frozen text-based reference anchor (e.g., the fixed semantic angles $\theta^{\text{text}}$), preserving a consistent semantic map across domains.
  }
  \label{fig:paradigm_comparison}
  \vspace{-3mm}
\end{figure}

The efficacy of the prevailing pre-training and fine-tuning paradigm faces challenges from real-world distributional shifts, which violate the independent and identically distributed (i.i.d.) assumption and cause severe performance degradation. While traditional domain adaptation (DA) methods~\cite{liang2020we,yang2021exploiting} can transfer knowledge to a new target domain, it is often at the cost of catastrophically forgetting prior ones. To address this limitation, Domain Incremental Learning (DIL)~\cite{van2022three} aims to continuously learn from new domains while preserving knowledge to maintain robust performance across all encountered domains.

Existing DIL approaches can generally be categorized into two major paradigms. The Unified Space Paradigm, shown in \cref{fig:paradigm_comparison}(a), attempts to project features from all domains into a single, shared visual embedding space~\cite{jeeveswaran2024gradual,shi2023unified}. While promoting knowledge sharing, this strategy struggles with large domain shifts. Visual features may differ considerably—not only in appearance but even in their underlying semantic variations. In such cases, forcing visually inconsistent features to align with each other can cause semantic distortion, inter-class confusion, and indistinct decision boundaries. On the other hand, the Isolated Space Paradigm, depicted in \cref{fig:paradigm_comparison}(b), employs parameter-isolation techniques like prompts or adapters~\cite{xu2025componential,feng2024cp,wang2022s} to learn domain-specific parameters. Although this effectively prevents inter-domain interference, it leads to knowledge fragmentation, creating isolated knowledge islands that hinder leveraging shared class-aware semantics across domains and exacerbate forgetting.

This raises a critical question: \textbf{How can a model bridge these disparate domains to share knowledge without falling into the trap of direct feature unification?}  
We posit that the key lies in a more stable, invariant reference structure.
Language, as a high-density carrier of human prior knowledge, provides a semantic space whose conceptual relations remain consistent across visual domains~\cite{devillers2021does,radford2021learning}.
Several DA studies~\cite{diko2024laguna} have shown that aligning the relative structure between source and target visual spaces through a shared semantic space is more effective than attempting direct feature alignment.
The domain-invariant nature of the semantic space inspires our \textit{Language-as-an-Anchor} design, which serves as the foundation for constructing a consistent relative alignment space.

To resolve this trade-off between interference and fragmentation, we propose \textbf{LAVA} (\textbf{L}anguage-\textbf{A}nchored \textbf{V}isual \textbf{A}lignment), a framework that employs language as a stable anchor to guide the alignment of visual representations based on their relative geometry. 
LAVA actively aligns each domain’s relative visual geometry to the underlying semantic geometry while preserving domain-specific characteristics. This design constructs a consistent relative visual space that not only mitigates inter-domain interference but also facilitates cross-domain communication and knowledge reuse.
Specifically, we first introduce the \textbf{(a) Vision-Language Relative Structural Alignment (VL-RSA) module}. 
As conceptualized in \cref{fig:paradigm_comparison}(c), LAVA adopts the relative geometry of class semantics as a text-based reference anchor. This anchor is formed by computing the pairwise cosine similarity angles between all classes of embeddings, thus capturing their intrinsic geometric relationships. Similarly, for each new domain, we introduce a learnable visual anchor and compute the relative similarity relationships of each input feature in the same manner. A structural alignment loss then enforces that the resulting relational representation of incoming visual features mirrors the fixed geometry of the text-based anchor. 
This module forges a domain-invariant relational scaffold, acting as a semantic bridge that maintains consistent geometry and links knowledge across domains.
Building on this foundation, to further enhance feature discriminability, we propose the \textbf{(b) Class-Aware Cross-Domain Feature Aggregation (CA-CDFA)} module. By leveraging this shared relational structure, its attention mechanism bypasses the comparison of incompatible raw features. Instead, it computes meaningful cross-domain similarities, which directly facilitates robust knowledge reuse and integration to combat forgetting.
During inference, to ensure accurate parameter activation and boost final performance, we propose a Multi-Level Feature Integration (MLFI) module for Domain Identification.
In summary, our contributions are as follows:
\begin{itemize}
    \item We explore the fundamental DIL dilemma of inter-domain interference versus knowledge fragmentation by introducing LAVA, a framework that preserves relative visual geometry across domains via a language-anchored semantic bridge, enabling robust knowledge aggregation.
    \item We design the VL-RSA module to align textual and visual relative structures via a structural alignment loss, and the CA-CDFA module to retrieve prior domain features via similarity-based attention, enabling cross-domain knowledge reuse and class-aware feature aggregation.
    \item Extensive experiments on four benchmarks demonstrate that LAVA achieves state-of-the-art performance, validating both its effectiveness and generalizability.
\end{itemize}

%% file: sec/2_formatting.tex
\section{Related Work}
\subsection{Domain Incremental Learning}
Domain Incremental Learning (DIL) tackles catastrophic forgetting by enabling a model to learn from a sequence of domains with varying data distributions while retaining prior knowledge.
Conventional DIL methods can be broadly categorized into three classes: architecture-based~\cite{yoon2017lifelong,hung2019compacting}, regularization-based~\cite{li2017learning,hendrycks2018benchmarking}, and replay-based methods~\cite{jeeveswaran2024gradual,shi2023unified}. 
Recently, prompt-based learning~\cite{xu2025componential,liu2024compositional,feng2024cp} has shown promise by preserving domain-specific knowledge via learnable prompts. 
However, these methods often suffer from several limitations: they are heavily dependent on the initial domain quality~\cite{wang2024non}, highly sensitive to prompt insertion design~\cite{feng2024cp}, and poorly adaptable to large domain shifts~\cite{xu2025componential}.
In contrast, our method leverages domain-invariant language as an anchor to construct a cross-domain relative alignment space, facilitating flexible class-aware representation aggregation across domains, effectively alleviating the limitations.

\subsection{Language Guidance in Domain Adaptation}
Vision-language models such as CLIP~\cite{radford2021learning} demonstrate strong zero-shot transferability. However, fine-tuning on specific downstream tasks often degrades their generalization to unseen domains~\cite{wortsman2022robust,kumar2022fine}. To mitigate this, recent Domain Adaptation (DA) studies has explored leveraging textual information to bridge the domain gap \cite{dunlap2023using,gokhale2021attribute,huang2023sentence,min2022grounding,wang2024landa}. Among them, \cite{diko2024laguna} proposes an adaptation strategy based on the geometric structure of a language-invariant latent space, encouraging structural consistency across domains by aligning the classifier with the underlying semantic geometry.
Inspired by this, we extend the language guidance relative geometry alignment to the more challenging DIL setting, where the model must incrementally adapt to new domains while avoiding catastrophic forgetting. 

\subsection{Relative Encodings}
Recent studies have demonstrated that latent spaces derived from similarly trained networks, while typically misaligned in absolute terms, retain consistent internal geometric relationships~\cite{moschella2022relative}. Consequently, representing data points using relative encodings—vectors capturing cosine similarities between each data point and a predefined set of anchors—enables tasks such as zero-shot model stitching~\cite{moschella2022relative} and cross-model or cross-modal representation translation~\cite{maiorca2023latent,norelli2023asif}. Moreover, predefined invariances can be explicitly integrated into these representations to make them more expressive~\cite{cannistraci2023bricks}. Relative encodings have also been successfully employed in downstream tasks, such as action anticipation~\cite{diko2024semantically}. Building on this paradigm, LAVA leverages relative encodings within a domain-agnostic, text-based reference anchor to model semantic inter-class relationships and facilitate alignment across diverse domains.

%% file: sec/3_Preliminary_and_Method.tex
\section{Preliminary}
\subsection{Problem Formulation}
Domain Incremental Learning (DIL) focuses on training a unified model sequentially across a series of $T$ domains, denoted by $\mathcal{D} = \{\mathcal{D}_t\}_{t=1}^{T}$, where each domain $\mathcal{D}_t = (\mathcal{X}_t, \mathcal{Z}_t)$ consists of a labeled training set $\mathcal{X}_t$ and a corresponding test set $\mathcal{Z}_t$.
During the $t$-th training stage, the model has access only to the current training data $\mathcal{X}_t$ and is prohibited from accessing any previous training sets $\mathcal{X}_{1 \sim t-1} = \bigcup_{\tau=1}^{t-1} \mathcal{X}_\tau$. After completion of training in domain $t$, the model is evaluated in the union of all test sets seen so far: $\mathcal{Z}_{1 \sim t} = \bigcup_{\tau=1}^{t} \mathcal{Z}_\tau$.
Each training set $\mathcal{X}_t$ contains $N_t$ labeled examples, i.e., $\mathcal{X}_t = \{(x_{t,i}, y_{t,i})\}_{i=1}^{N_t}$, where $x_{t,i}$ is the $i$-th input image and $y_{t,i}$ is its associated label. The test set $\mathcal{Z}_t$ follows the same structure. Let $\mathcal{C}_t$ denote the label set of domain $t$, where $\forall i \in [1, N_t],\, y_{t,i} \in \mathcal{C}_t$. 
All domains share a common label space $\mathcal{C}$ with $N_c$ classes, i.e., $\mathcal{C}_1 = \mathcal{C}_2 = \dots = \mathcal{C}_T = \mathcal{C}$.

\subsection{Prompt-based Learning}
\label{sec:prompt}
Following the prompt design in S-Prompts~\cite{wang2022s}, we keep the visual encoder $f_\theta$ frozen and maintain a pool of prompts 
$\mathcal{P} = \{\mathbf{P}_{(t)}\}_{t=1}^{T}$. Each prompt $\mathbf{P}_{(t)} \in \mathbb{R}^{N_p \times D}$ serves as a lightweight adaptor for domain $t$, where $N_p$ is the prompt length and $D$ matches the hidden dimension of the visual encoder.
Given an image $x_i$ from domain $t$, its corresponding prompt $\mathbf{P}_{(t)}$ is inserted between the patch embeddings $\mathbf{x}_{\text{emb}}$ and the class token $\mathbf{x}_{\text{cls}}$, 
forming an extended sequence
$\mathbf{x}_{p} = [\mathbf{x}_{\text{cls}},\, \mathbf{P}_{(t)},\, \mathbf{x}_{\text{emb}}]$, 
which is then processed by the frozen backbone to yield the visual representation:
\begin{equation}
\mathbf{g}_i = f_\theta(\mathbf{x}_{p}).
\label{eq:prompt}
\end{equation}
This parameter-efficient and modular design allows the model to capture domain-specific knowledge while mitigating catastrophic forgetting, 
enabling continual adaptation without modifying the shared backbone.

\subsection{Text-based Reference Anchor}
Following \cite{diko2024laguna}, we use a pre-trained text encoder $G$ to generate a set of semantic anchors, $\mathcal{A}^{\text{Text}} \in \mathbb{R}^{N_c \times D_l}$, by encoding the class names of all $N_c$ categories.
To capture the fixed geometric structure of the semantic space, we compute a \textit{relative encoding} for each class anchor. Let $v$ be a specific class anchor, corresponding to one of the rows in $\mathcal{A}^{\text{Text}}$. Its relative encoding $r^v$ is computed by measuring its cosine similarity to all anchors in $\mathcal{A}^{\text{Text}}$:
\begin{equation}
r^v = \text{rel}(v, \mathcal{A}^{\text{Text}}) = [\cos(v, \mathcal{A}^{\text{Text}}[1]), \dots, \cos(v, \mathcal{A}^{\text{Text}}[N_c])],
\label{eq:r_v}
\end{equation}
where $\cos(\cdot, \cdot)$ denotes the cosine similarity. The resulting vectors form a static reference structure of inter-class relationships. During training, the model is encouraged to align its relative visual geometry with this reference, thereby preserving consistent class relationships across domains.

\begin{figure*}[t]   
  \centering
  \includegraphics[width=0.95\textwidth]{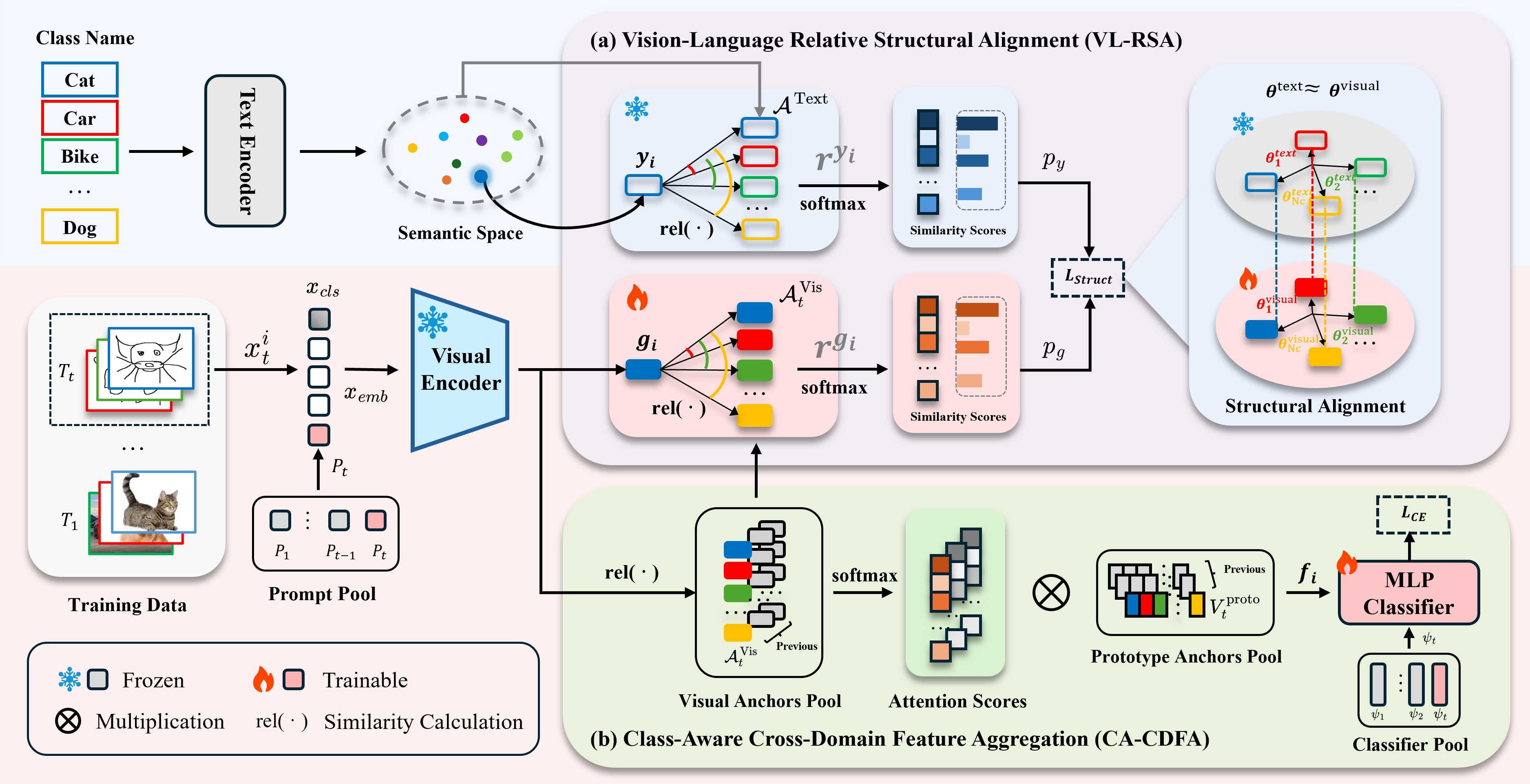} 
  \caption{\textbf{Overview of the proposed LAVA framework.} 
LAVA consists of two branches: the text branch establishes a text-based reference anchor from class names and computes relative encodings that capture their pairwise relationships; the visual branch employs domain-specific prompts to extract features from a frozen encoder. These two branches feed into two core modules: (a) The VL-RSA module aligns the visual representations with the text-based geometry via a structural alignment loss, creating a domain-invariant relational space; (b) The CA-CDFA module aggregates class-aware features from previously seen domains using an attention mechanism, enabling effective cross-domain knowledge reuse and aggregation.
}
\label{fig:framework}
\vspace{-3mm}
\end{figure*}

\section{Method}
\subsection{Overall Framework}
\Cref{fig:framework} illustrates the overall framework of our proposed LAVA.   
Prior to training, we use a pre-trained text encoder to embed all class names, thereby establishing a static semantic reference space that captures their inter-class geometric relationships.
To effectively capture domain-specific characteristics, we augment the frozen visual encoder with a set of learnable prompts tailored to each domain.
Based on these prompt-adapted features, we first employ the Vision-Language Relative Structural Alignment (VL-RSA) module to align the relative structures of the visual and textual feature spaces. Subsequently, the Class-Aware Cross-Domain Feature Aggregation (CA-CDFA) module aggregates relevant features from previously encountered domains to facilitate effective knowledge reuse.

\subsection{Vision-Language Relative Alignment}
\label{sec:structure_alignment}
After obtaining the prompt-adapted visual representations (\cref{sec:prompt}), 
we aim to align their relational structure with a fixed semantic geometry anchored by language. To achieve this, we introduce a learnable \emph{Visual Anchor} matrix $\mathcal{A}^{\text{Vis}}_{(t)} \in \mathbb{R}^{N_c \times D_v}$ for each domain $t$, where $N_c$ is the number of classes and $D_v$ is the output dimension of the visual encoder $f_\theta$. All domain-specific \emph{Visual Anchors} collectively form a \emph{Visual Anchor pool} $\mathcal{A}^{\text{Vis}} = \{\mathcal{A}^{\text{Vis}}_{(1)}, \dots, \mathcal{A}^{\text{Vis}}_{(T)}\}$, where $T$ is the total number of domains. 
Similar to the text-based anchor $\mathcal{A}^{\text{Text}} \in \mathbb{R}^{N_c \times D_l}$, each  $\mathcal{A}^{\text{Vis}}_{(t)}$ maintains a class-wise structure with one anchor per class.
We enforce consistency between their respective \textit{relative encodings}: the input feature's similarity to visual anchors is aligned with its corresponding class name embedding's similarity to the text anchors. 
Specifically, the textual \textit{relative encoding} is computed from the pre-defined \emph{Text-based Anchor} $\mathcal{A}^{\text{Text}}$ (see \cref{eq:r_v}). The semantic reference encoding corresponding to the ground-truth label $y_i$ is defined as:
\begin{equation}
\mathbf{r}^{y_i} = \operatorname{rel}(\mathcal{A}^{\text{Text}}[y_i], \mathcal{A}^{\text{Text}}).
\label{eq:ryi}
\end{equation}
Since the pre-trained text encoder $G$ is frozen, $\mathbf{r}^{y_i}$ remains fixed during training.
Then, the prompt-adapted feature $\mathbf{g_i}$ is also mapped into a \textit{relative encoding} using cosine similarity with respect to $\mathcal{A}^{\text{Vis}}_{(t)}$:
\begin{equation}
\mathbf{r}^{g_i} = \operatorname{rel}(\mathbf{g_i}, \mathcal{A}_{(t)}^{\text{Vis}}).
\label{eq:rgi}
\end{equation}
We encourage the alignment between $\mathbf{r}^{g_i}$ and $\mathbf{r}^{y_i}$ by minimizing the Kullback–Leibler divergence between their softmax-normalized distributions:
\begin{equation}
\mathcal{L}_{\text{Struct}} = \mathrm{KL}(p_g \,\|\, p_y) = \sum_{c=1}^{N_c} p_g^{(c)} \log \frac{p_g^{(c)}}{p_y^{(c)}},
\label{eq:kl}
\end{equation}
where $p_g = \operatorname{softmax}(\mathbf{r}^{g_i})$ and $p_y = \operatorname{softmax}(\mathbf{r}^{y_i})$.

\subsection{Cross-Domain Feature Aggregation}
\label{sec:CA-CDFA}
Once the \emph{Visual Anchors} are structurally aligned with the \emph{Text-based Anchor}, the resulting visual geometry space becomes \emph{comparable across domains}. We exploit this relative representation to facilitate effective class-wise aggregation of cross-domain features.

For each domain, we also construct a domain-specific learnable \emph{Prototype Anchor} \( V^{\text{proto}}_{(t)} \in \mathbb{R}^{N_c \times D_v} \), where each row represents the latent semantic center of a class in domain \( t \). Collectively, these anchors form a \emph{Prototype Anchor Pool}
\(\mathcal{V}^{\text{proto}} = \{ V^{\text{proto}}_{(1)}, V^{\text{proto}}_{(2)}, \dots, V^{\text{proto}}_{(T)} \}\),
which encapsulates class-level semantic representations across all observed domains.
Unlike the structure-aligned \emph{Visual Anchors} \( \mathcal{A}^{\text{Vis}}_{(t)} \), which emphasize geometry consistency across domains, these \emph{Prototype Anchors} are designed to capture domain-specific class semantics, thereby enhancing the model's discriminative capability. Each \( V^{\text{proto}}_{(t)} \) is trained only on data from its corresponding domain.

We perform a key–value attention mechanism, using the input feature as a query, the \emph{Visual Anchors} as keys, and the \emph{Prototype Anchors} as values. This design enables attention computation over structure-aligned representations (keys), while retrieving domain-aware class semantics (values).
Specifically, given an input feature \( \mathbf{g}_i\), we first concatenate all \emph{Visual Anchors} into a global key set: 
\( \mathcal{A}^{\text{global}}_{(t)} = \bigl\Vert_{\tau=1}^{t}\mathcal{A}^{\text{Vis}}_{(\tau)} \in \mathbb{R}^{tN_c \times D_v} \). Then, we compute the relative encoding of \( \mathbf{g}_i \) with respect to this global key set:
\begin{equation}
\tilde{\alpha}_i = \operatorname{softmax}\bigl(\operatorname{rel}(\mathbf{g}_i, \mathcal{A}^{\text{global}}_{(t)})\bigr) \in \mathbb{R}^{tN_c},
\label{eq:global_alpha}
\end{equation}
yielding attention scores over class-level anchors across all domains simultaneously.
The corresponding global value set is constructed by concatenating all domain-specific \emph{Prototype Anchors}:
\(V_{(t)}^{\text{global}} = \bigl\Vert_{\tau=1}^{t} V^{\text{proto}}_{(\tau)} \in \mathbb{R}^{tN_c \times D_v}\).
Finally, the aggregated feature $\mathbf{f}_i$ is obtained via attention-weighted aggregation followed by a residual connection:
\begin{equation}
\mathbf{f}_i = \tilde{\alpha}_i\,V_{(t)}^{\text{global}} + \mathbf{g}_i.
\label{eq:global_fused}
\end{equation}

\noindent\textbf{Training Objective.}
The fused feature \( \mathbf{f}_i \) is passed to the current domain classifier \( \psi_{(t)} \) to produce the prediction \( \hat{y}_i = \psi_{(t)}(\mathbf{f}_i) \). The overall training objective combines cross-entropy loss with structural alignment loss:
\begin{equation}
\mathcal{L}
   \;=\;
   \mathcal{L}_{\text{CE}}\!\bigl(\hat{y}_i, y_i\bigr)
   \;+\;
   \lambda\,\mathcal{L}_{\text{Struct}}.
\label{eq:combine_loss}
\end{equation}
where $\mathcal{L}_{\text{CE}}$ denotes the cross-entropy loss, and $\lambda$ is a hyperparameter that balance the contributions of each term. 

During training, all domain-specific prompts, anchors, and classifiers are randomly initialized for each domain and frozen after completion, 
ensuring stable knowledge accumulation without cross-domain interference.

\subsection{Inference via Multi-Level Feature Integration}
\label{sec:MLFI}
\begin{figure}[t]   
  \centering
  \includegraphics[width=0.45\textwidth]{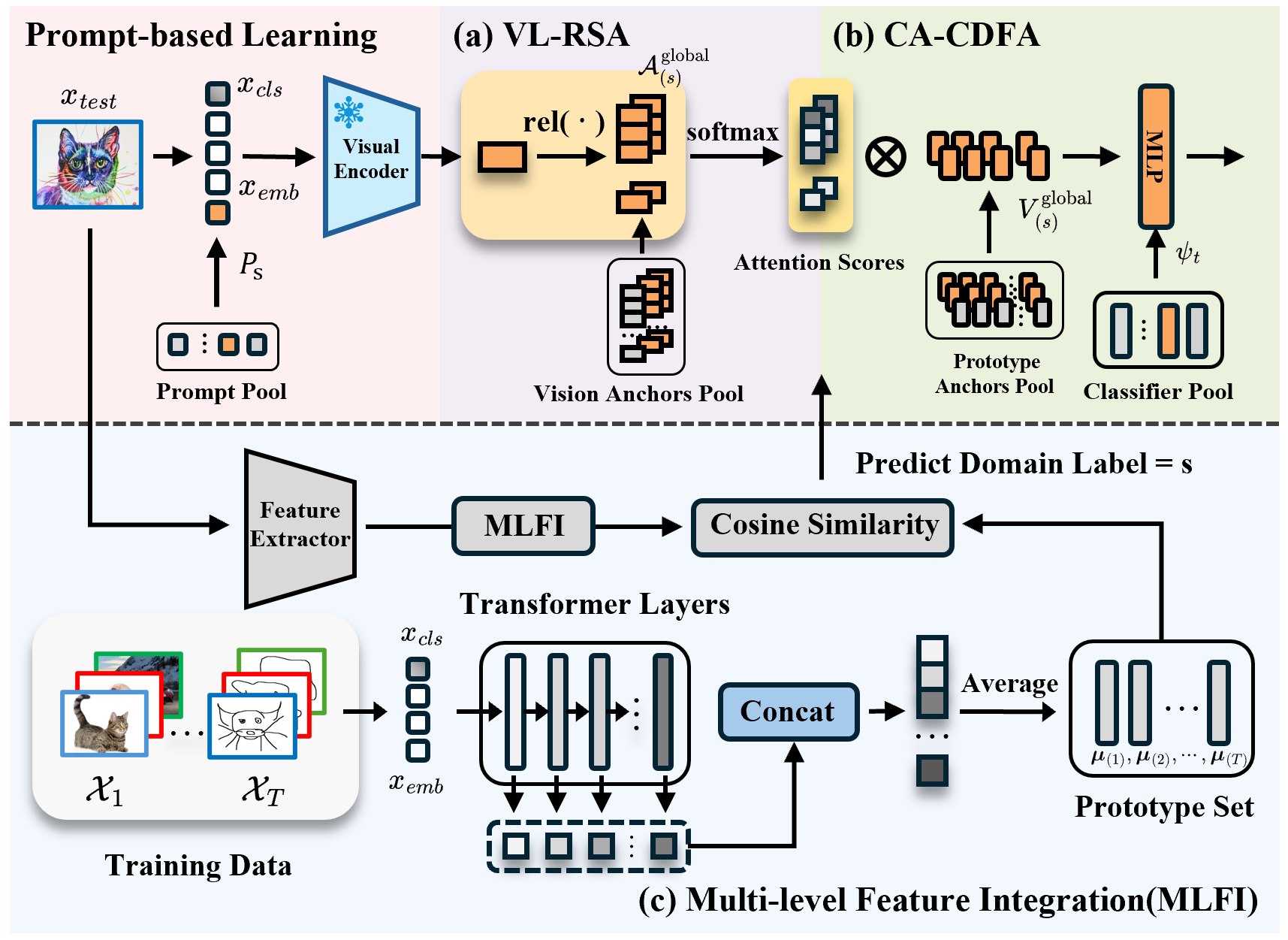}
\caption{\textbf{Overview of the inference pipeline.} The model first identifies the domain of an input image via (c) Multi-Level Feature Integration (MLFI) module (bottom), and then employs the domain-specific modules (top) for the final classification task.}
\label{fig:inference_flow}
\vspace{-5mm}
\end{figure}

In the inference phase, the model first identifies the domain of a test image. As shown in \cref{fig:inference_flow}, we extract its feature $F(x_{\text{test}})$ and compare it against the pre-computed domain prototypes $\mathcal{M}$. The domain whose prototype has the highest cosine similarity is selected, and the corresponding parameters are activated for final classification.

To facilitate accurate domain identification, we introduce the Multi-Level Feature Integration (MLFI) module. The key intuition is that different layers of a visual encoder capture complementary information (\cref{fig:layer_ablation}), ranging from low-level textures to high-level semantics. MLFI aggregates this information to form a more discriminative representation, thereby producing more reliable domain prototypes. Specifically, we extract intermediate features $f^{(l)}(x)$ from a fixed set of layers $L$ (detailed in \cref{sec:MLFI_ablation}), and concatenate them into a unified multi-scale feature:
\begin{equation}
    F(x) = \text{Concat}\left( \left[ f^{(l)}(x) \right]_{l \in L} \right).
\label{eq:concat}
\end{equation}
For a given domain $t$, its prototype $\boldsymbol{\mu}_{(t)}$ is computed by averaging the multi-level features $F(x_i)$ of all its $N_t$ training samples, irrespective of their class labels:
\begin{equation}
    \boldsymbol{\mu}_{(t)} = \frac{1}{N_t} \sum_{i=1}^{N_t} F(x_i).
\label{eq:prototype}
\end{equation}
Repeating this process for all $T$ domains yields the final prototype set $\mathcal{M} = \{\boldsymbol{\mu}_{(1)}, \dots, \boldsymbol{\mu}_{(T)}\}$, with all prototypes computed offline after finishing training on each domain.

%% file: sec/4_Experiments.tex
\section{Experiments}

\begin{table*}[t]
  \centering
  \setlength{\tabcolsep}{3pt}
  \footnotesize
  \begin{tabular}{ll|ccc|ccc|ccc|ccc|ccc}
    \toprule
    \multirow{2}{*}{Methods} & \multirow{2}{*}{Publication}
    & \multicolumn{3}{c|}{DomainNet} 
    & \multicolumn{3}{c|}{ImageNet-R} 
    & \multicolumn{3}{c|}{ImageNet-C} 
    & \multicolumn{3}{c|}{ImageNet-Mix} 
    & \multicolumn{3}{c}{Average} \\
    \cmidrule(lr){3-5}
    \cmidrule(lr){6-8}
    \cmidrule(lr){9-11}
    \cmidrule(lr){12-14}
    \cmidrule(lr){15-17}
    & & $A_T\uparrow$ & $A_A\uparrow$ & $F_T\downarrow$
      & $A_T\uparrow$ & $A_A\uparrow$ & $F_T\downarrow$
      & $A_T\uparrow$ & $A_A\uparrow$ & $F_T\downarrow$
      & $A_T\uparrow$ & $A_A\uparrow$ & $F_T\downarrow$
      & $A_T\uparrow$ & $A_A\uparrow$ & $F_T\downarrow$ \\
    \midrule
    EWC~\cite{kirkpatrick2017overcoming}            & \scriptsize{\textcolor{darkgray}{\textit{NAS 2017}}}          & 55.04 & 51.31 & 27.52 & 47.05 & 45.50 & 21.30 & 45.07 & 45.07 & 26.69 & 42.22 & 43.46 & 24.75 & 47.35 & 46.34 & 25.07 \\
    LwF~\cite{li2017learning}            & \scriptsize{\textcolor{darkgray}{\textit{T-PAMI 2017}}}       & 59.01 & 55.02 & 16.84 & 54.47 & 52.89 & 14.23 & 42.87 & 42.87 & 27.72 & 50.06 & 51.54 & 19.13  & 51.60 & 50.58 & 19.48 \\
    L2P~\cite{wang2022learning}            & \scriptsize{\textcolor{darkgray}{\textit{CVPR 2022}}}         & 58.94 & 55.96 & 13.35 & 65.27 & 65.27 & 14.93 & 55.62 & 55.62 & 17.35 & 50.34 & 44.31 & 19.72 & 57.54 & 55.29 & 16.34 \\
    DualPrompt~\cite{wang2022dualprompt}     & \scriptsize{\textcolor{darkgray}{\textit{ECCV 2022}}}         & 62.68 & 60.63 & 13.83 & 71.10 & 67.92 & 7.19 & \underline{77.25} & \underline{77.25} & 2.51 & 71.23 & 73.32 & 7.17 & 70.57 & 69.78 & 7.68 \\
    CODA-Prompt~\cite{smith2023coda}    & \scriptsize{\textcolor{darkgray}{\textit{CVPR 2023}}}         & 56.73 & 52.87 & 23.88 & 69.07 & 66.25 & 14.12 & 52.30 & 52.30 & 25.15 & 59.02 & 52.82 & 20.22 & 59.28 & 56.06 & 20.85 \\
    CPrompt~\cite{gao2024consistent}       & \scriptsize{\textcolor{darkgray}{\textit{CVPR 2024}}}         & 60.18 & 60.62 & 1.19 & 74.02 & 75.27 & -0.70 & 49.71 & 49.71 & \textbf{-0.82} & 56.86 & 55.04 & \underline{-0.83} & 60.19 & 60.16 & \underline{-0.29} \\
    \midrule
    S-Prompts~\cite{wang2022s}     & \scriptsize{\textcolor{darkgray}{\textit{NeurIPS 2022}}}    & 67.51 & 66.40 & \underline{0.68} & 41.23 & 41.00 & 2.83 & 45.26 & 45.26 & 0.47 & 43.84 & 43.57 & 0.47 & 49.46 & 49.06 & 1.33 \\
    PINA~\cite{wang2024non}          & \scriptsize{\textcolor{darkgray}{\textit{ECCV 2024}}}       & 73.38 & 74.18 & 1.13 & 63.80 & 61.41 & 2.40 & 68.05 & 68.05 & 0.15 & 65.19 & 65.38 & 0.84 & 67.61 & 67.26 & 1.13 \\
    CP-Prompt~\cite{feng2024cp}    & \scriptsize{\textcolor{darkgray}{\textit{MM 2024}}}         & 72.16 & 73.11 & 0.95 & 60.98 & 59.79 & 2.35 & 63.91 & 63.91 & 0.30 & 61.46 & 61.54 & 1.13 & 64.63 & 64.59 & 1.18 \\
    C-Prompt*~\cite{liu2024compositional} & \scriptsize{\textcolor{darkgray}{\textit{IJCV 2024}}}       & 65.03 & 63.10 & 1.63 & 74.57 & 70.35 & \textbf{-9.56} & 69.89 & 69.89 & 2.49 & 67.90 & 66.43 & \textbf{-1.29} & 69.35 & 67.44 & \textbf{-1.68} \\
    KA-Prompt*~\cite{xu2025componential} & \scriptsize{\textcolor{darkgray}{\textit{ICML 2025}}}       & 67.06 & 65.24 & 2.32 & 75.12 & 71.85 & \underline{-5.83} & 74.12 & 74.12 & 3.97 &  71.27 & 71.65 & 0.17 & 71.89 & 70.72 & 0.16 \\
    \midrule
    \rowcolor{gray!10} LAVA-share (Ours) & \scriptsize{\textcolor{darkgray}{\textit{This Paper}}} & \underline{73.40} & \underline{74.64} & \underline{0.68} & \underline{80.27} & \underline{75.10} & -0.18 & 77.00 & 77.00 & 0.11 & \underline{77.73} & \underline{75.72} & 0.24 & \underline{77.00} & \underline{75.62} & 0.21 \\
    \rowcolor{gray!10} \textbf{LAVA (Ours)} & \scriptsize{\textcolor{darkgray}{\textit{This Paper}}} & \textbf{73.84} & \textbf{75.13} & \textbf{0.58} & \textbf{80.89} & \textbf{75.58} & -0.34 & \textbf{77.58} & \textbf{77.58} & \underline{0.04} & \textbf{78.45} & \textbf{76.56} & 0.26 & \textbf{77.69} & \textbf{76.21} & 0.14 \\
    \bottomrule
  \end{tabular}
  \caption{Comparison across four DIL benchmarks. We mark the best results in \textbf{bold} and second-best with \underline{underlines}. (*) C-Prompt and KA-Prompt employ specific mechanisms that can result in negative forgetting (i.e., knowledge gain). The proposed \textit{LAVA-share} denotes a lightweight variant of LAVA that reuses the anchor pool as both keys and values.}
  \label{tab:all_results}
  \vspace{-3mm}
\end{table*}

\begin{figure*}[t]
  \centering
  \begin{subfigure}[b]{0.46\textwidth}
    \includegraphics[width=\textwidth]{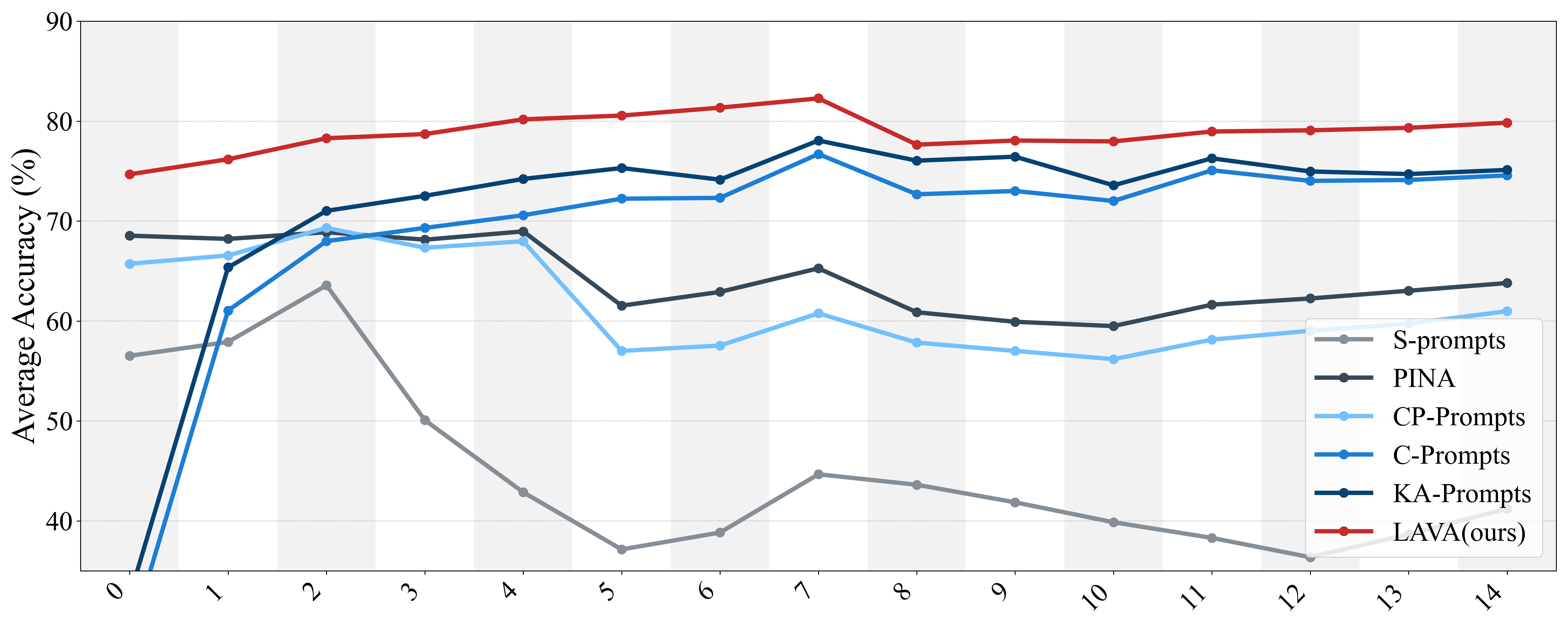}
    \caption{ImageNet-R average accuracy on seen domains}
    \label{fig:imagenet-r}
    \vspace{-2mm}
  \end{subfigure}
  \hspace{0.02\textwidth}  
  \begin{subfigure}[b]{0.46\textwidth}
    \includegraphics[width=\textwidth]{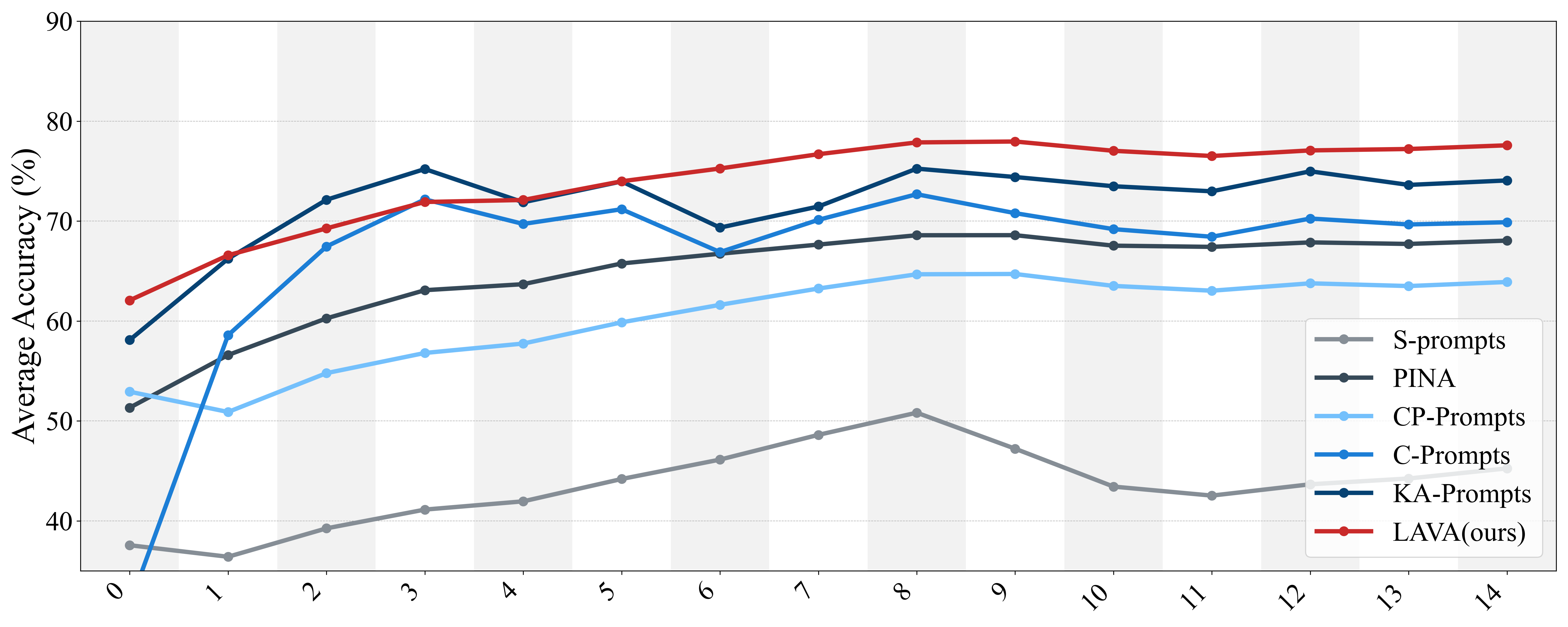}
    \caption{ImageNet-C average accuracy on seen domains}
    \label{fig:imagenet-c}
    \vspace{-2mm}
  \end{subfigure}
\caption{Performance progression on the ImageNet-R and ImageNet-C benchmarks.}
  \label{fig:seen-domain-trend}
  \vspace{-5mm}
\end{figure*}

\minisection{Datasets.}
Following~\cite{xu2025componential}, our experiments are conducted on four DIL benchmarks: DomainNet~\cite{peng2019moment}, ImageNet-R~\cite{hendrycks2021many}, ImageNet-C~\cite{hendrycks2018benchmarking} and ImageNet-Mix~\cite{xu2025componential}.
Specifically, DomainNet provides 6 domains with 345 classes. 
ImageNet-R consists of 200 classes across 15 style domains. 
For ImageNet-C, we use the setup of~\cite{xu2025componential}, using 200 classes (identical to ImageNet-R) across 15 corruption types as domains.
ImageNet-Mix is a fusion of ImageNet-R and ImageNet-C, comprising 30 domains overall.

\minisection{Evaluation Metrics.}
We evaluate our method using three standard DIL metrics: 
(1) Average Accuracy ($A_A$), the overall accuracy computed over all test samples from all domains;
(2) Average Task Accuracy ($A_T$), the mean of task accuracies, measuring performance balance;
(3) Forgetting Degree ($F_T$), a measure of catastrophic forgetting, following Core50~\cite{lomonaco2017core50}, where lower values are better.
We also report average performance across four DIL benchmarks. See Appendix \cref{sec:Evaluation_Metrics} for metric definitions.

\minisection{Implementation Details.}
We employ the AdamW optimizer ($\beta_1=0.9$, $\beta_2=0.999$) with an initial learning rate of 0.01 and a cosine decay schedule.
Additionally, we apply a weight decay of $10^{-4}$ for regularization to mitigate overfitting.
The default training epochs are set to 50.
The length of prompt token $N_p$ is set to 16.
The mini-batch size is set to 128. 
The training images are resized to 224$\times$224.
A temperature parameter $\tau=0.07$ is applied in the softmax normalization to scale the cosine-similarity scores when constructing relational distributions.
All experiments are conducted on a single NVIDIA RTX 4090 GPU.

\minisection{Baselines.}
We benchmark LAVA against state-of-the-art methods from both Domain- and Class-Incremental Learning (DIL and CIL). 
DIL baselines include prompt-based methods (S-Prompts~\cite{wang2022s}, CP-Prompt~\cite{feng2024cp}, C-Prompt~\cite{liu2024compositional}, KA-Prompt~\cite{xu2025componential}) and the adapter-based method PINA~\cite{wang2024non}. 
Among these, three methods rely on explicit domain identification at inference to select their domain-specific parameters: S-Prompts and CP-Prompt use a KNN-based strategy, whereas PINA adopts its dedicated Patch Shuffle Selector (PSS) module.
We follow the configurations described in their respective papers (see Appendix \cref{sec:Domain_Identification}). 
Ablations on this module are further discussed in \cref{sec:MLFI_ablation}.
For CIL, we consider regularization-based methods (LwF~\cite{li2017learning}, EWC~\cite{kirkpatrick2017overcoming}) and prompt-based ones (L2P~\cite{wang2022learning}, DualPrompt~\cite{wang2022dualprompt}, CODA-Prompt~\cite{smith2023coda}, CPrompt~\cite{gao2024consistent}).
All methods are implemented on a \textbf{CLIP (ViT-B/16)} backbone using official or verified re-implementations, and evaluated under the \textbf{rehearsal-free} setting without storing past samples.

\subsection{Main Results}

\begin{table*}[t]
  \centering
  \footnotesize
  \setlength{\tabcolsep}{3pt}
  \begin{tabular}{cccc|ccc|ccc|ccc|ccc}
    \toprule
    \multirow{2}{*}{Baseline} & \multirow{2}{*}{VL-RSA} & \multirow{2}{*}{CA-CDFA} & \multirow{2}{*}{MLFI} 
    & \multicolumn{3}{c|}{DomainNet} 
    & \multicolumn{3}{c|}{ImageNet-R} 
    & \multicolumn{3}{c|}{ImageNet-C} 
    & \multicolumn{3}{c}{ImageNet-Mix} \\
    \cmidrule(lr){5-7} \cmidrule(lr){8-10} \cmidrule(lr){11-13} \cmidrule(lr){14-16}
    & & & 
    & $A_T\uparrow$ & $A_A\uparrow$ & $F_T\downarrow$ 
    & $A_T\uparrow$ & $A_A\uparrow$ & $F_T\downarrow$ 
    & $A_T\uparrow$ & $A_A\uparrow$ & $F_T\downarrow$ 
    & $A_T\uparrow$ & $A_A\uparrow$ & $F_T\downarrow$ \\
    \midrule
    \checkmark &            &            &            & 67.51 & 66.40 & 0.68 & 41.23 & 41.00 & 2.83 & 45.26 & 45.26 & 0.47 & 43.84 & 43.57 & 0.47 \\
    \checkmark & \checkmark &            &            & 73.36 & 74.38 & 1.13 & 78.50 & 73.84 & 0.81 & 75.77 & 75.77 & 0.17 & 75.72 & 73.15 & 0.66 \\
    \checkmark & \checkmark & \checkmark &            & 73.65 & 74.86 & 1.01 & 79.92 & 74.90 & 0.08 & 77.17 & 77.17 & 0.24 & 77.74 & 75.71 & 0.45 \\
    \midrule
    \checkmark & \checkmark & \checkmark & \checkmark & \textbf{73.84} & \textbf{75.13} & \textbf{0.58} & \textbf{80.89} & \textbf{75.58} & \textbf{-0.34} & \textbf{77.58} & \textbf{77.58} & \textbf{0.04} & \textbf{78.45} & \textbf{76.56} & \textbf{0.26} \\
    \bottomrule
  \end{tabular}
\caption{Component contributions. The baseline corresponds to S-Prompts with prompt tuning on a frozen backbone, 
a unified classifier and a KNN-based domain identification strategy.  Even without MLFI, LAVA achieves state-of-the-art performance under this setting.}

  \label{tab:ablation}
  \vspace{-4mm}
\end{table*}

\begin{table}[t]
\centering
\footnotesize
\setlength{\tabcolsep}{5pt}
\begin{tabular}{@{}llccc ccc@{}}
\toprule
\multicolumn{2}{c}{} & \multicolumn{3}{c}{\textbf{DomainNet}} & \multicolumn{3}{c}{\textbf{ImageNet-R}} \\
\cmidrule(lr){3-5} \cmidrule(lr){6-8}
 & & $A_T{\uparrow}$ & $A_A{\uparrow}$ & $F_T{\downarrow}$ & $A_T{\uparrow}$ & $A_A{\uparrow}$ & $F_T{\downarrow}$ \\
\midrule
 & random & 73.04 & 73.95 & 1.59 & 77.06 & 72.47 & 1.11 \\
\midrule
\multirow{2}{*}{ST} & IDs & 72.98 & 73.94 & 1.34 & 80.22 & 75.34 & -0.07 \\
 & class names & 73.44 & 74.45 & 1.32 & 80.37 & 75.54 & 0.30 \\
\midrule
\multirow{2}{*}{CLIP} & IDs & 73.19 & 74.14 & 1.23 & 79.32 & 74.51 & 0.27 \\
 & class names & \textbf{73.84} & \textbf{75.13} & \textbf{0.58} & \textbf{80.89} & \textbf{75.58} & \textbf{-0.34} \\
\bottomrule
\end{tabular}
\caption{Ablation on alignment reference structures. “ST” and “CLIP” denote text encoders (Sentence Transformer and CLIP); “IDs” and “class names” indicate the semantic input type.}
\label{tab:reference_structures}
\vspace{-2mm}
\end{table}

\minisection{Comparison with State-of-the-Arts.}
Our LAVA model demonstrates a significant leap over existing state-of-the-art methods, as detailed in \cref{tab:all_results}. It achieves a new benchmark with an average Task Accuracy ($A_T$) of \textbf{77.69\%} and Average Accuracy ($A_A$) of \textbf{76.21\%}, outperforming the previous SOTA, KA-Prompt, by \textbf{5.80\%} and \textbf{5.49\%} respectively. 
LAVA's superiority is most pronounced on benchmarks with severe domain shifts, such as DomainNet (\textbf{+6.78\%}) and ImageNet-Mix (\textbf{+7.18\%}), directly validating the robustness of our relative alignment strategy. 
Beyond conventional accuracy metrics, LAVA demonstrates strong knowledge retention, achieving a relatively low forgetting degree ($F_T$) of \textbf{0.14}. While specialized methods like KA-Prompt and C-Prompt employ explicit reuse mechanisms to obtain numerically lower forgetting, LAVA remains one of the more effective approaches among standard baselines in mitigating knowledge degradation.
We further introduce a lightweight variant, \textbf{LAVA-share}, which eliminates the separate \emph{Prototype Anchor Pool} and instead reuses the \emph{Visual Anchor Pool} as both keys and values in the CA-CDFA module. 
In this way, a single anchor pool simultaneously serves as the key–value pair, 
achieving a \textbf{32\%} parameter reduction without sacrificing performance (see \cref{minisec:acc_vs_params}).
In addition, the learning process visualizations on ImageNet-R and ImageNet-C (\cref{fig:seen-domain-trend}) further confirm LAVA’s consistent superiority, particularly in the more challenging domains encountered in the later stages of training.

\begin{figure}[t]
    \centering
    \includegraphics[width=0.45\textwidth]{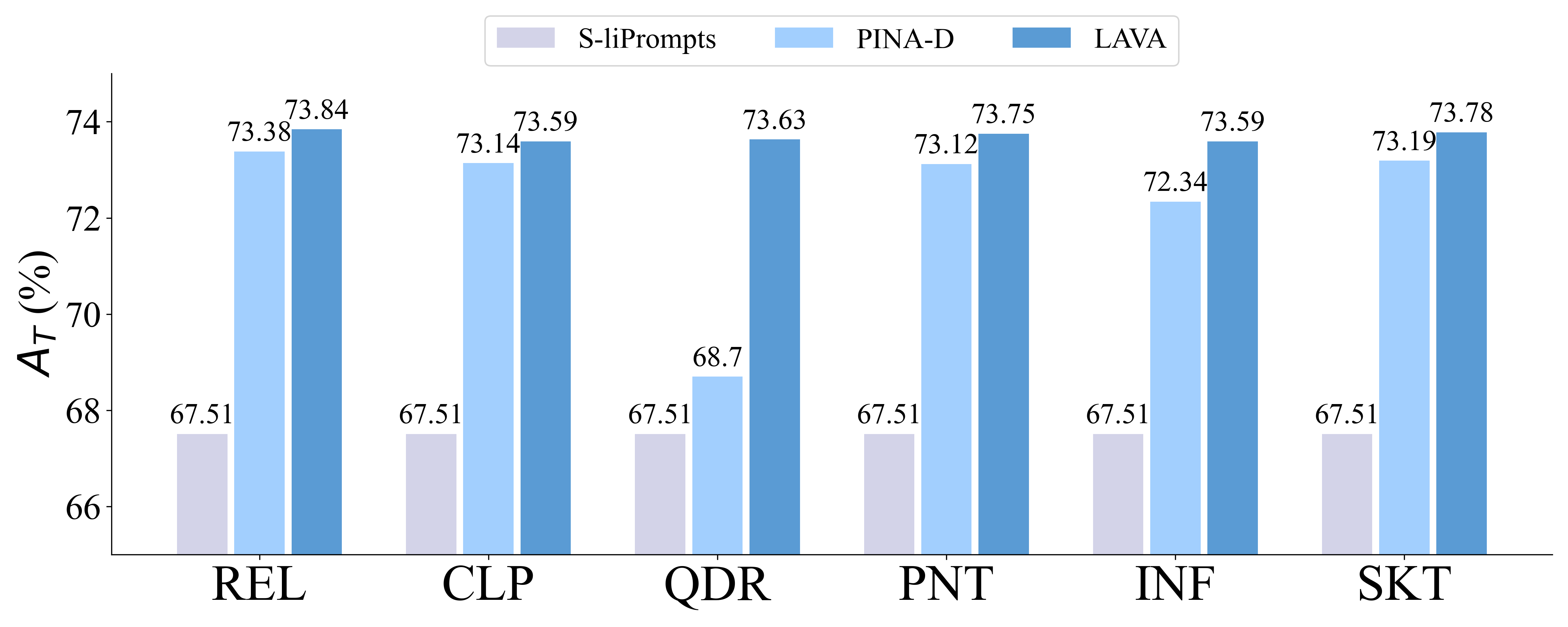}
\caption{Impact of Domain Order on DomainNet. LAVA shows robust stability and performance across domain orders, including the challenging \textit{Quickdraw}-first order.}
\label{fig:domain_order}
\vspace{-5mm}
\end{figure}

\subsection{Ablation Study}
\minisection{Component Ablations.} 
\Cref{tab:ablation} summarizes our component ablation study. 
We use S-Prompts~\cite{wang2022s} combined with a standard KNN-based domain identification strategy as our baseline.
On top of this baseline, we incrementally add three core components of LAVA:
(1) Vision-Language Relative Structural Alignment (VL-RSA), which yields a substantial performance gain by enforcing relative structural alignment between vision and language, validating our relative alignment strategy;
(2) Class-Aware Cross-Domain Feature Aggregation (CA-CDFA), which significantly improves cross-domain knowledge reuse and enhances feature discriminability, effectively overcoming knowledge fragmentation;
(3) Multi-Level Feature Integration (MLFI), which provides a modest yet consistent improvement by facilitating more accurate domain identification at inference. Notably, even without this module, our method can still achieve state-of-the-art performance using the standard KNN-based domain identification approach.

\begin{table}[t]
\centering
\footnotesize
\setlength{\tabcolsep}{5pt}
\begin{tabular}{@{}l ccc ccc@{}}
\toprule
 & \multicolumn{3}{c}{\textbf{DomainNet}} & \multicolumn{3}{c}{\textbf{ImageNet-Mix}} \\
\cmidrule(lr){2-4} \cmidrule(lr){5-7}
\textbf{Method} & {$A_T{\uparrow}$} & {$A_A{\uparrow}$} & {$F_T{\downarrow}$} &
{$A_T{\uparrow}$} & {$A_A{\uparrow}$} & {$F_T{\downarrow}$} \\
\midrule
w/o $L_{\text{Struct}}$ & 73.55 & 74.56 & 1.03 & 77.10 & 75.39 & 0.61 \\
L1                      & 73.12 & 74.12 & 0.62 & 76.80 & 74.71 & 0.37 \\
L2                      & 73.45 & 74.54 & 0.68 & 77.52 & 75.35 & 0.32 \\
KL                      & \textbf{73.84} & \textbf{75.13} & \textbf{0.58} & \textbf{78.45} & \textbf{76.56} & \textbf{0.26} \\
\bottomrule
\end{tabular}
\caption{Ablation of structure alignment loss formulations, including $\ell_1$, $\ell_2$, KL divergence, and a variant without $L_{\text{Struct}}$.}
\label{tab:struct_loss}
\vspace{-4mm}
\end{table}

\minisection{Impact of Domain Order.}
We evaluate LAVA's robustness to domain order on DomainNet using various permutations (see Appendix \cref{sec:appendix_domain_order}). As shown in \cref{fig:domain_order}, LAVA maintains top performance with minimal fluctuation, while PINA drops significantly on the \textit{Quickdraw}-first sequence~\cite{wang2024non}, where LAVA surpasses it by \textbf{4.93\%} in $A_T$.

\minisection{Ablation on Alignment Reference and Objectives.}
We jointly analyze the effects of different reference structures and alignment objectives. 
As shown in \cref{tab:reference_structures}, we compare random initialization, numerical class IDs, and class names encoded by a Sentence Transformer (ST)~\cite{diko2024laguna} or CLIP’s text encoder. 
Using CLIP-encoded class names yields the best results, confirming that its pre-trained semantic geometry is naturally aligned with the visual space. 
We further evaluate several formulations of the structure alignment loss $L_{\text{Struct}}$ (\cref{tab:struct_loss}), including $\ell_1$, $\ell_2$, KL divergence, and a variant without alignment. 
The KL objective achieves the best accuracy, indicating that penalizing distributional discrepancies enables more stable and discriminative alignment.

\begin{figure}[t]
    \centering
    \begin{minipage}[t]{0.48\linewidth}
        \centering
        \includegraphics[width=\linewidth]{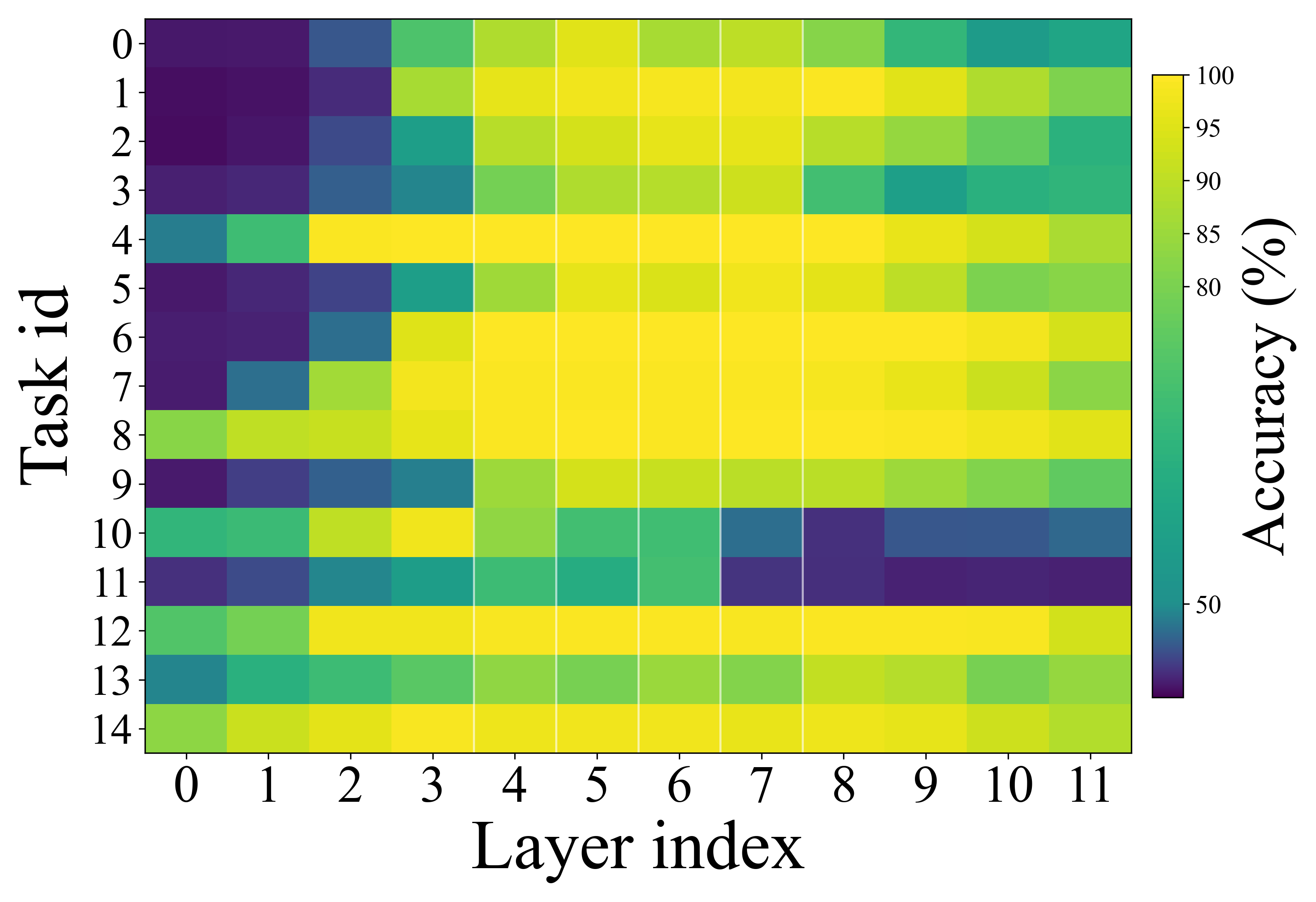}
        \subcaption{}
        \vspace{-3mm}
    \end{minipage}
    \hfill
    \begin{minipage}[t]{0.48\linewidth}
        \centering
        \includegraphics[width=\linewidth]{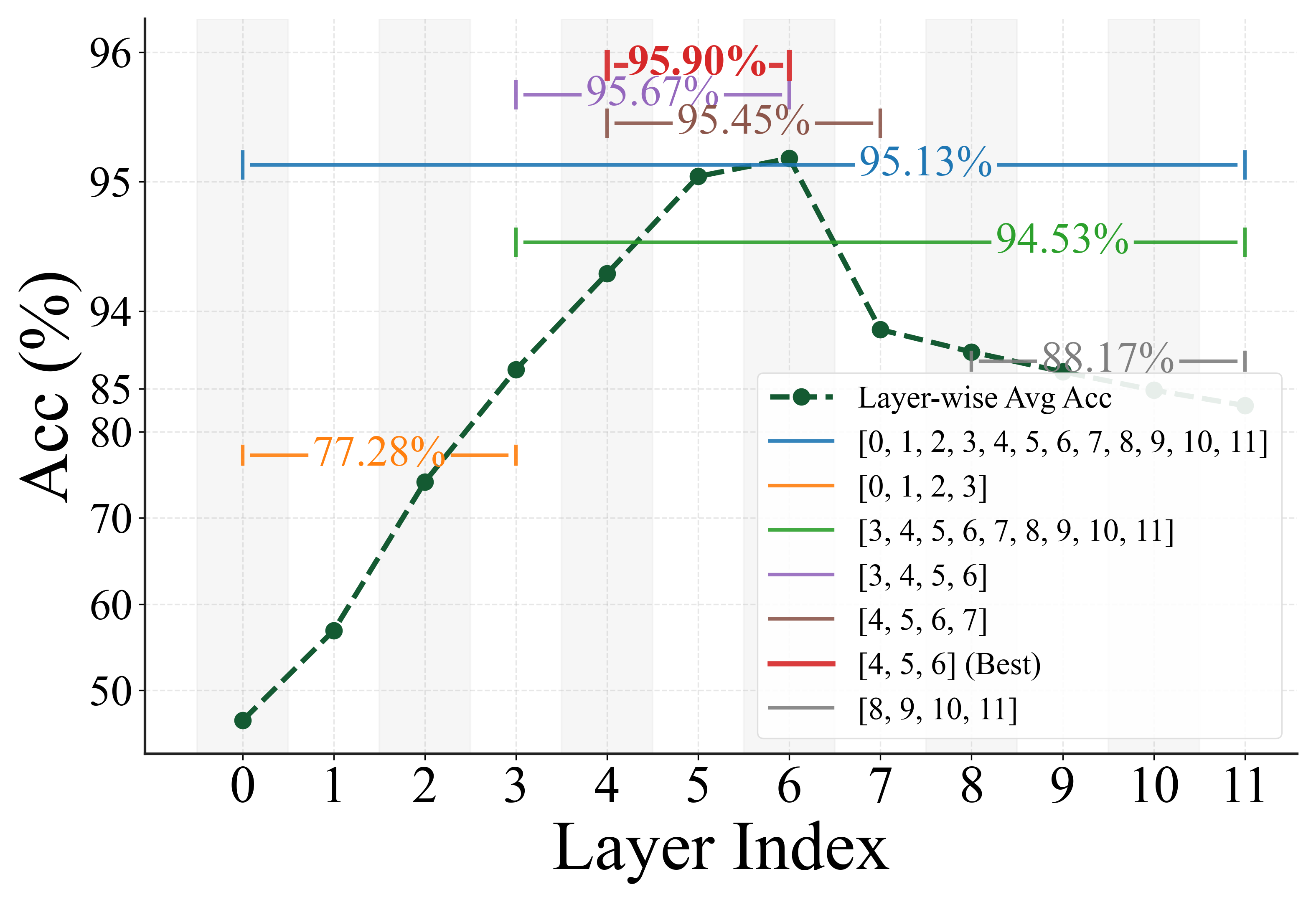}
        \subcaption{}
        \vspace{-3mm}
    \end{minipage}
\caption{
Ablation on layer selection in the MLFI module on ImageNet-C. 
(a) Per-layer accuracy across all tasks, illustrating the contribution of each layer. 
(b) Mean per-layer accuracy (green dashed) and results of different layer-integration combinations.
}
    \vspace{-3mm}
    \label{fig:layer_ablation}
\end{figure}

\minisection{Ablation on Layer Selection in MLFI.}
\label{sec:MLFI_ablation}
We conduct a systematic ablation to validate the design of the Multi-Level Feature Integration (MLFI) module.
\Cref{fig:layer_ablation}(a) presents the per-task accuracy using features from individual layers on ImageNet-C, while (b) shows the average accuracy of each layer and different layer-integration combinations across tasks.
The results indicate that different layers encode complementary visual cues—some performing well on certain tasks but less effectively on others.
By integrating multiple layers, MLFI exploits these complementarities to achieve more consistent and robust domain identification.
Notably, combining layers $L^{*} = \{4, 5, 6\}$ yields the highest accuracy on ImageNet-C, surpassing both single-layer and other integration settings.
Further details on the layer-selection search space and the associated memory overhead are provided in Appendix \cref{sec:best_layer_searching}.
Moreover, \Cref{tab:multi_dataset_results} compares MLFI with representative approaches~\cite{wang2024non}, including K-Nearest Neighbors (K-NN), Nearest Mean Classifier (NMC) and Patch Shuffle Selector (PSS) (see Appendix \cref{sec:Domain_Identification}). MLFI’s superior domain-identification accuracy consistently translates into higher overall performance.

\sisetup{detect-weight=true, detect-inline-weight=math}

\begin{table}[t]
\centering
\footnotesize
\setlength{\tabcolsep}{4pt}
\renewcommand{\arraystretch}{1.15}
\resizebox{\columnwidth}{!}{%
\begin{tabular}{@{}lcccccccc@{}}
\toprule
 & \multicolumn{2}{c}{\textbf{DomainNet}} 
 & \multicolumn{2}{c}{\textbf{ImageNet-R}} 
 & \multicolumn{2}{c}{\textbf{ImageNet-C}} 
 & \multicolumn{2}{c}{\textbf{ImageNet-Mix}} \\
\cmidrule(lr){2-3} \cmidrule(lr){4-5} \cmidrule(lr){6-7} \cmidrule(lr){8-9}
\textbf{Method} 
& {$A_{\text{cls}}{\uparrow}$} & {$A_A{\uparrow}$}
& {$A_{\text{cls}}{\uparrow}$} & {$A_A{\uparrow}$}
& {$A_{\text{cls}}{\uparrow}$} & {$A_A{\uparrow}$}
& {$A_{\text{cls}}{\uparrow}$} & {$A_A{\uparrow}$} \\
\midrule
K-NN  & 84.03 & 74.66 & 56.12 & 74.90 & 82.72 & \underline{77.17} & 77.52 & \underline{75.71} \\
NMC   & 85.02 & 74.76 & \underline{57.46} & \underline{75.47} & \underline{83.25} & 77.12 & \underline{77.74} & 75.70 \\
PSS   & \underline{85.07} & \underline{74.78} & 53.43 & 73.76 & 74.02 & 75.03 & 64.11 & 72.29 \\
\midrule
MLFI  & \textbf{86.45} & \textbf{75.13} & \textbf{57.88} & \textbf{75.58} & \textbf{95.90} & \textbf{77.58} & \textbf{88.00} & \textbf{76.56} \\
\bottomrule
\end{tabular}
}
\vspace{-2mm}
\caption{Comparison of domain identification strategies. $A_{\text{cls}}$ is the domain identification accuracy, while $A_A$ is the final task Average Accuracy achieved using that method.
}
\label{tab:multi_dataset_results}
\vspace{-2mm}
\end{table}

\begin{figure}[t]
    \centering
    \begin{minipage}[t]{0.48\linewidth}
        \centering
        \includegraphics[width=\linewidth]{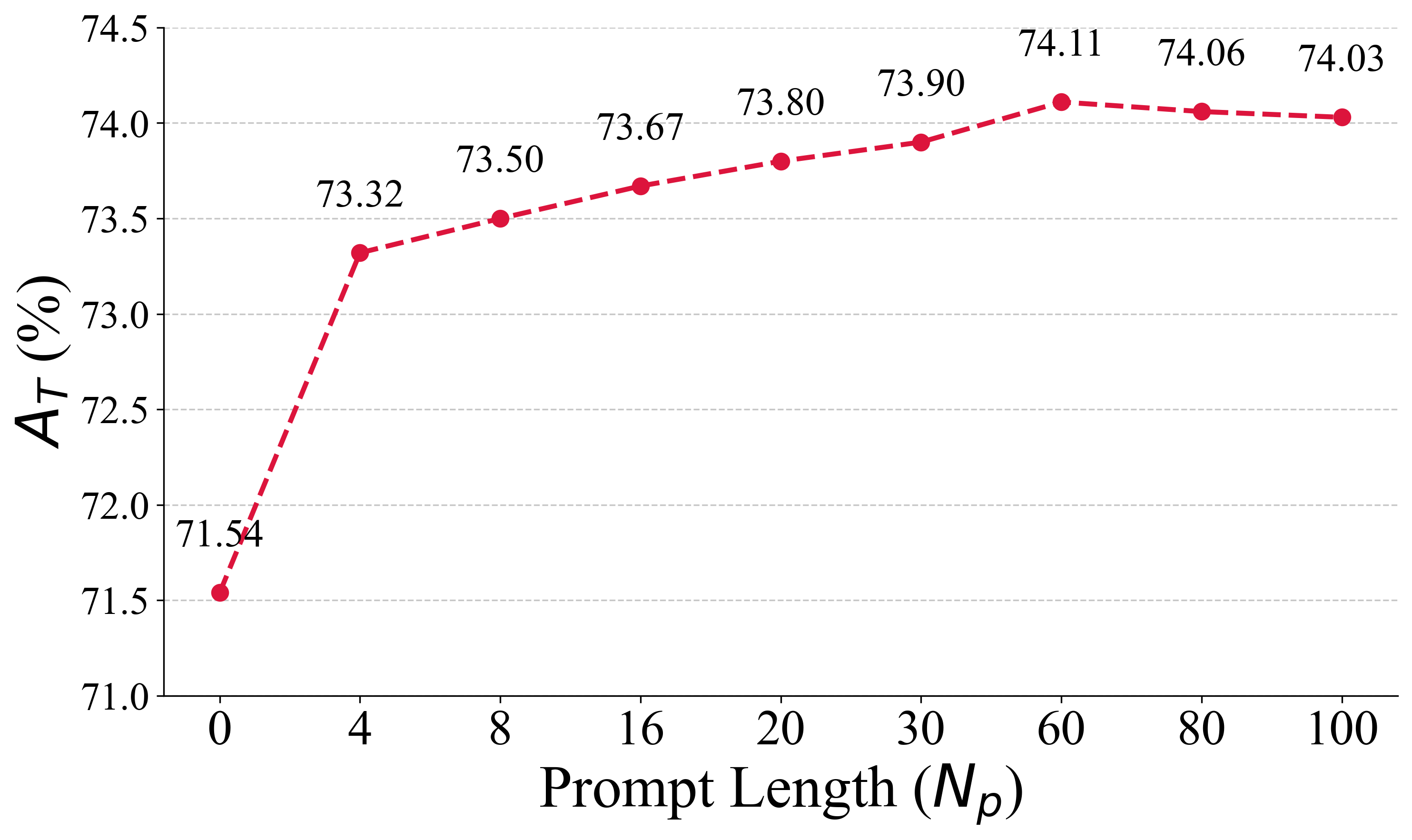}
        \subcaption{}
        \vspace{-3mm}
    \end{minipage}
    \hfill
    \begin{minipage}[t]{0.48\linewidth}
        \centering
        \includegraphics[width=\linewidth]{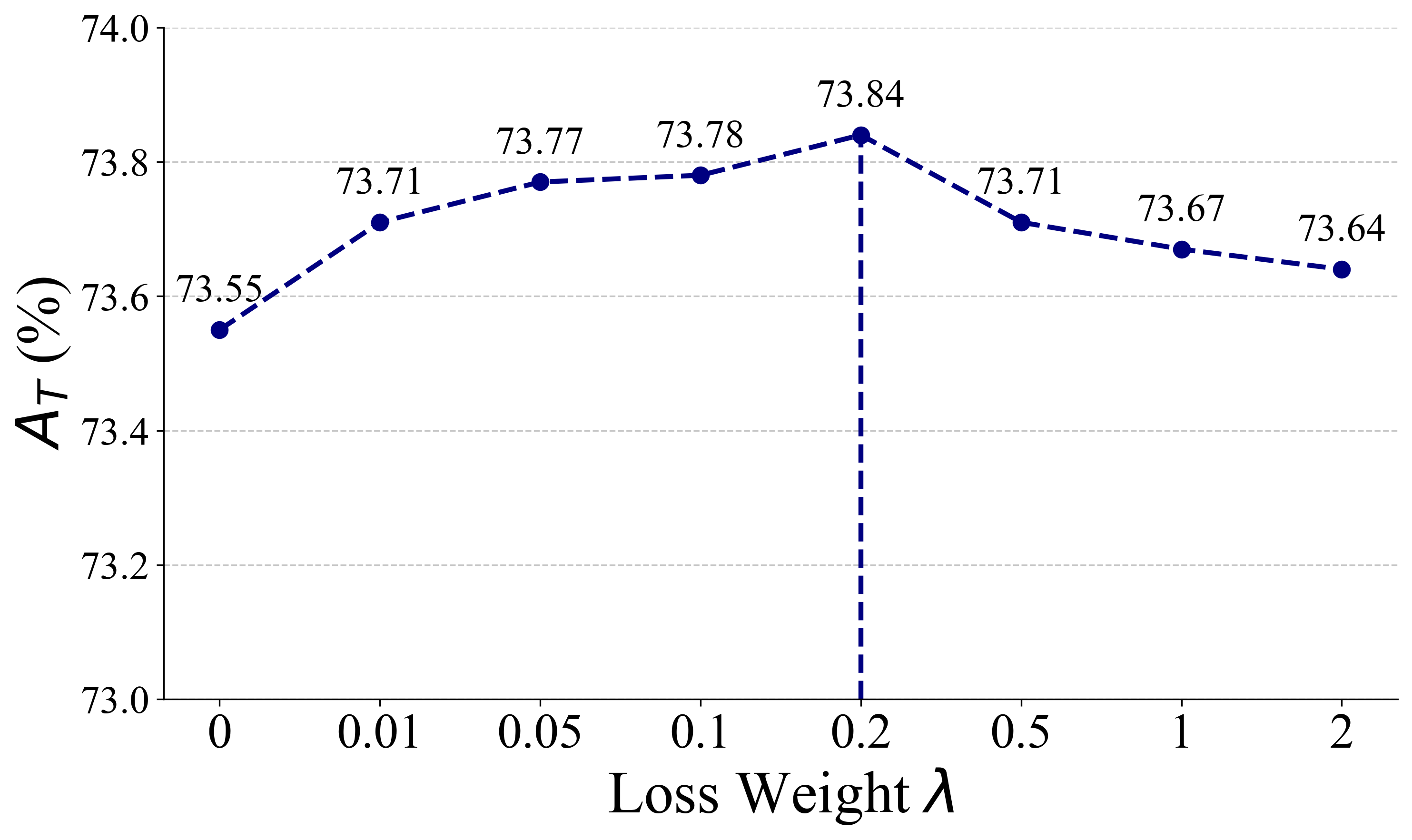}                  
        \subcaption{}
        \vspace{-3mm}
    \end{minipage}
    \caption{Hyperparameter sensitivity study on the DomainNet dataset. (a) Prompt length $N_p$. (b) Loss weight $\lambda$.}
    \label{fig:hyperparam}
    \vspace{-3mm}
\vspace{-1mm}
\end{figure}

\subsection{Further Analysis}
\minisection{Hyperparameter Sensitivity Analysis.}
As shown in \cref{fig:hyperparam}, we analyze LAVA’s sensitivity to two key hyperparameters on DomainNet: the prompt length $N_p$ and the alignment loss weight $\lambda$.
The model remains largely robust to the choice of $N_p$, with performance improving only marginally as the prompt length increases. For consistency with prior methods, we set the prompt length to $N_p{=}16$.
For $\lambda$, the optimal value differs slightly across datasets but lies within a similar range; for DomainNet we use $\lambda{=}0.2$. Full settings are listed in Appendix \cref{sec:lambda}.

\minisection{Parameter–Performance Trade-off.}
\label{minisec:acc_vs_params}
\Cref{fig:acc_vs_params} compares accuracy versus trainable parameters on ImageNet-R (detailed in Appendix \cref{sec:parameter_comparison}).
LAVA achieves state-of-the-art accuracy with high parameter efficiency, requiring only 7.05M trainable parameters—\textbf{33\% fewer} than KA-Prompt (the previous SOTA)—while attaining higher task accuracy.
The lightweight variant, \textit{LAVA-share}, further reduces the parameter count to 4.75M yet continues to outperform all competing approaches by a substantial margin.
These results indicate that LAVA’s advantage arises from its architectural efficiency rather than increased model capacity. 

\begin{figure}[t]
    \centering
    \includegraphics[width=0.43\textwidth]{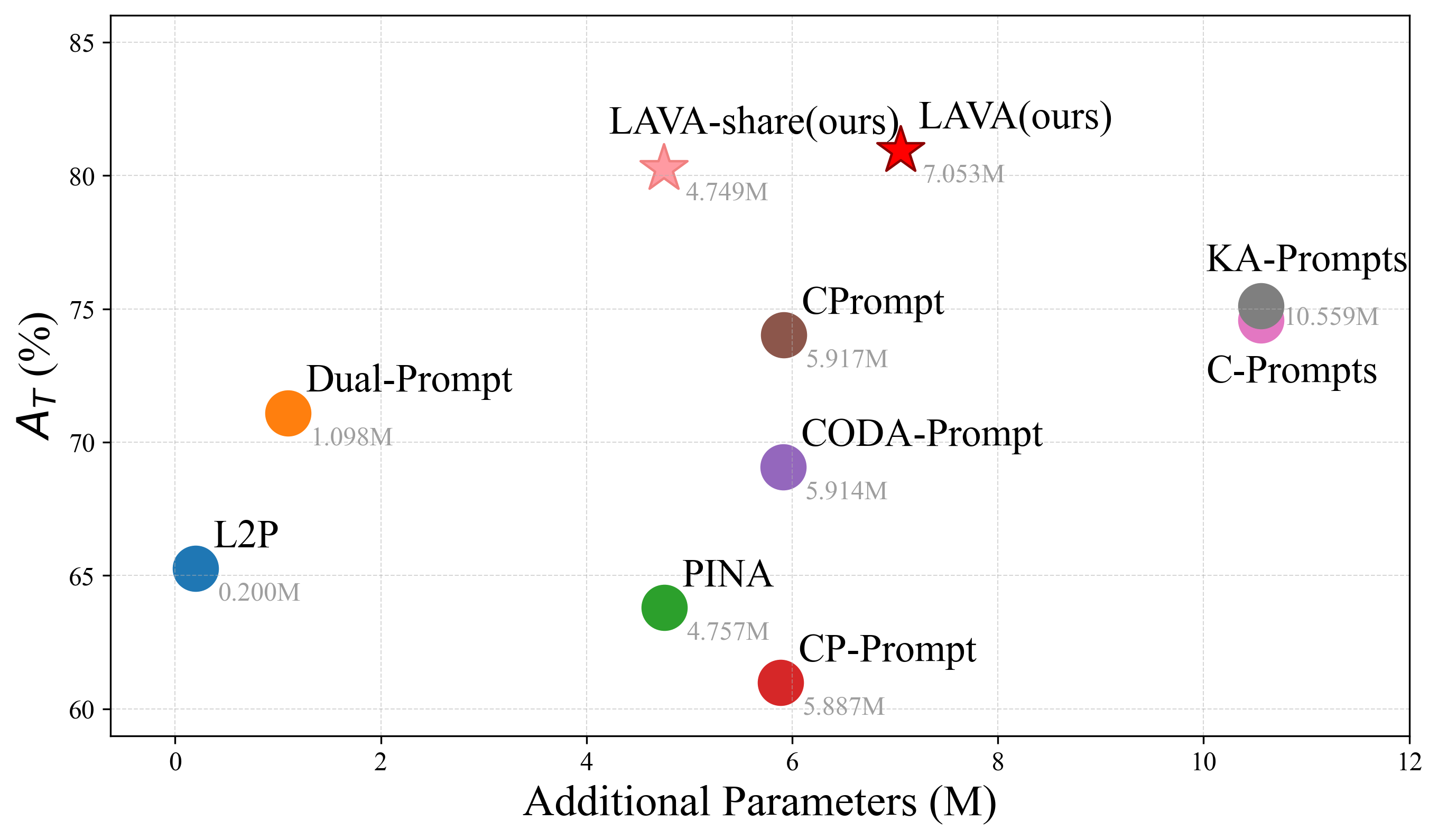}
    \vspace{-3mm}
    \caption{Performance–Parameter comparison on ImageNet-R.}
    \label{fig:acc_vs_params}
    \vspace{-1mm}
\end{figure}

\begin{table}[t]
  \centering
  \footnotesize
  \setlength{\tabcolsep}{2pt}
  \resizebox{0.9\columnwidth}{!}{%
    \begin{tabular}{c c c c}
      \toprule
      \textbf{T} & \textbf{Cumulative (M, +\%)} & \textbf{Active@Inf. (M, +\%)} & \textbf{GPU (GB)} \\
      \midrule
      1  & 0.47 (+0.31\%)  & \multirow{4}{*}{\textbf{150.092 (+0.31\%)}} & \multirow{4}{*}{\textbf{2.49}} \\
      5  & 2.35 (+1.57\%)  &  &  \\
      10 & 4.70 (+3.14\%)  &  &  \\
      15 & 7.05 (+4.71\%)  &  &  \\
      \bottomrule
    \end{tabular}%
  }
  \caption{Scalability across domains. Percentages are measured w.r.t. the backbone (149.62M).}
  \label{tab:scalability}
  \vspace{-5mm}
\end{table}

\minisection{Model Scalability.}
As shown in \cref{tab:scalability}, the trainable parameters grow linearly with the number of domains, reaching 7.05M at $T{=}15$ only—4.7\% of the 149.6M backbone.
At inference, however, LAVA activates only a fixed set of parameters (150.09M in total), introducing merely \textbf{$\sim$0.3\%} additional parameters beyond the backbone and keeping the GPU memory footprint constant at 2.49~GB. 
This indicates that domain expansion increases only the stored model size, while the inference-time cost remains lightweight and unchanged. 
A detailed analysis of LAVA’s computational overhead and efficiency is provided in Appendix \cref{sec:parameter_comparison}.
\vspace{-2mm}

%% file: sec/5_Conclusion.tex
\section{Conclusion and Future Works}
In this work, we explore the fundamental dilemma in DIL—the trade-off between inter-domain interference and knowledge fragmentation. 
We propose LAVA, a novel framework that preserves relative visual geometry across domains via a language-anchored semantic bridge, enabling effective retrieval of prior knowledge and robust feature aggregation. 
Extensive experiments on multiple DIL benchmarks demonstrate that LAVA consistently outperforms state-of-the-art methods. 
Beyond  empirical success, LAVA introduces the idea of leveraging an external, stable, and knowledge-rich structure—such as language—as a semantic compass to navigate the drift of internal representations, offering a promising path toward more stable and scalable domain incremental systems.
We believe that aligning relative geometry opens promising directions for broader applications, such as few-shot and multi-modal learning.

\vspace{0.5ex}
\noindent\textbf{Acknowledgements.}
This work is supported in part by the National Natural Science Foundation of China under Grant 62476054, and in part by the Fundamental Research Funds for the Central Universities of China. This research work is supported by the Big Data Computing Center of Southeast University.

%% file: sec/X_suppl.tex
\clearpage
\setcounter{page}{1}
\maketitlesupplementary

\section{Supplementary}
\subsection{Details of the Evaluation Metrics}
\label{sec:Evaluation_Metrics}
To provide a comprehensive assessment, our evaluation framework is built upon three core metrics, each chosen to target a distinct pillar of continual learning performance: overall effectiveness, balanced performance across domains, and resistance to catastrophic forgetting.

\minisection{Average Accuracy ($A_A$)}
This metric measures the overall classification performance across all test samples from all domains after the model has been fully trained on all $T$ domains. It is defined as:
\[
A_A = \frac{\sum_{i=1}^T c_{T,i}}{\sum_{i=1}^T |\mathcal{Z}_i|}
\]
where $c_{T,i}$ denotes the number of correctly classified samples in the test set $\mathcal{Z}_i$ by the final model $M_T$, and $|\mathcal{Z}_i|$ is the total number of samples in that test set.

\minisection{Average Task Accuracy ($A_T$)}  
This metric emphasizes balanced performance across domains, computing the mean of per-domain accuracies. It is calculated as:
\[
A_T = \frac{1}{T}\sum_{i=1}^T a_{T,i}
\]
where $a_{T,i}$ is the accuracy of the final model $M_T$ on the test set of the $i$-th domain, $\mathcal{Z}_i$.

\minisection{Forgetting Degree ($F_T$)}  
This metric quantifies catastrophic forgetting by measuring how performance on previously learned domains changes as new domains are introduced. It is calculated by averaging the per-domain backward transfer ($BWT_i$) across all previously learned domains:
\[
BWT_i = \frac{1}{T-i} \sum_{j=i+1}^{T} (a_{j,i} - a_{i,i})
\]
The overall Forgetting Degree is then defined as:
\[
F_T = \frac{1}{T-1} \sum_{i=1}^{T-1} BWT_i
\]
Here, $a_{j,i}$ denotes the accuracy on domain $i$ after training up to domain $j$, and $a_{i,i}$ is the accuracy on domain $i$ immediately after it was learned. 

Since model performance on previous domains typically decreases as new domains are learned (i.e., $a_{j,i} < a_{i,i}$), the raw forgetting degree defined above is generally negative. For clearer interpretation, our tables report its negation while still using the symbol $F_T$. Under this convention, larger positive values indicate stronger forgetting, whereas smaller values reflect better knowledge retention. Notably, when $F_T$ becomes negative, it implies positive backward transfer, meaning that learning later domains unexpectedly improves performance on earlier ones.

\subsection{Details of Domain Identification Strategies}
\label{sec:Domain_Identification}

Several baselines in our comparison rely on explicit domain identification during inference.
For example, S-Prompts \cite{wang2022s} and CP-Prompts \cite{feng2024cp} employ a K-NN-based strategy, whereas PINA \cite{wang2024non} adopts its dedicated Patch Shuffle Selector (PSS).
For completeness, we summarize these strategies below.

\minisection{K-Nearest Neighbors (K-NN).}
These methods use the frozen visual encoder to extract features for each training domain and apply K-Means clustering to obtain a small set of domain prototypes.
During inference, a test sample is encoded with the same frozen encoder and matched to the nearest prototype (typically via L2 distance); the identified domain determines which domain-specific parameters are used for inference.

\minisection{Patch Shuffle Selector (PSS).}
PSS \cite{wang2024non} constructs domain prototypes by applying random patch shuffling to each training image before feature extraction, suppressing class-dependent structure while retaining domain-level appearance statistics.
The prototype for each domain is computed as the mean feature of its shuffled samples.
At inference, the test sample is encoded without shuffling and assigned to the nearest prototype using cosine similarity; the identified domain determines the domain-specific parameters used for inference.

\minisection{Nearest Mean Classifier (NMC).}
Consistent with PINA \cite{wang2024non}, we also include NMC as a comparison method.
NMC computes a single mean feature for each domain directly from the frozen-encoder embeddings of all training samples, without clustering or patch shuffling.
During inference, the test sample is assigned to the domain whose mean prototype yields the highest cosine similarity.

As shown in \cref{tab:multi_dataset_results}, we compare these three strategies—K-NN, PSS, and NMC—under a unified evaluation protocol.
For all baselines, we follow the configurations reported in their respective papers.

\begin{table*}[t]
\centering
\begin{tabular}{c|cccccc}
\hline
Order & 1st & 2nd & 3rd & 4th & 5th & 6th \\
\hline
1 & real & quickdraw & painting & sketch & infograph & clipart \\
2 & clipart & infograph & painting & quickdraw & real & sketch \\
3 & quickdraw & infograph & clipart & painting & sketch & real \\
4 & painting & sketch & clipart & real & infograph & quickdraw \\
5 & infograph & clipart & painting & quickdraw & real & sketch \\
6 & sketch & clipart & painting & quickdraw & real & infograph \\
\hline
\end{tabular}
\caption{The six domain order permutations for the DomainNet dataset used in our experiments.}
\label{tab:domain_orders}
\end{table*}

\subsection{Domain Order Permutations on DomainNet}
\label{sec:appendix_domain_order}

To rigorously evaluate the robustness of our model, LAVA, against variations in domain order, we constructed six distinct domain permutations for the DomainNet dataset.

The first sequence follows the methodology of \cite{xu2025componential}, where domains are sorted by decreasing data size to simulate a challenging Domain Incremental Learning (DIL) scenario. The second sequence adopts the default DomainNet order, which is consistent with the experimental setups in several baseline methods \cite{wang2022s,feng2024cp,wang2024non}.

The design of the remaining four permutations is primarily motivated by the evaluation strategy in PINA~\cite{wang2024non}. Acknowledging the significant domain gaps among the six domains in DomainNet (\textit{Real}, \textit{Quickdraw}, \textit{Painting}, \textit{Sketch}, \textit{Infograph}, and \textit{Clipart}), the initial domain can substantially impact subsequent model adaptation. Therefore, mirroring PINA's setup, we ensure that each of the six domains serves as the starting point in one of the permutations.

However, to create a more demanding continual learning testbed, we introduce an additional layer of complexity. While PINA's setting focuses on varying the initial domain, we further randomize the sequence of the subsequent domains. Specifically, after designating a unique starting domain for each permutation, we also randomly shuffle the order of the remaining five domains. This approach not only varies the initial domain exposure but also alters the entire sequence of domain transitions, enabling a more comprehensive and challenging assessment of model robustness.

The six resulting domain order permutations are detailed in \cref{tab:domain_orders}, and their impact on model performance is illustrated in \cref{fig:domain_order}.

\begin{table}[t]
  \centering
  \footnotesize
  \setlength{\tabcolsep}{10pt}
  \begin{tabular}{l c}
    \toprule
    \textbf{Dataset} & \textbf{Selected Layers} \\
    \midrule
    DomainNet      & $\{9, 10, 11\}$ \\
    ImageNet-R     & $\{10, 11\}$    \\
    ImageNet-C     & $\{4, 5, 6\}$   \\
    ImageNet-Mix   & $\{6, 11\}$     \\
    \bottomrule
  \end{tabular}
  \caption{Selected MLFI layers for each benchmark.}
  \label{tab:mlfi_layers}
\end{table}

\begin{table}[t]
  \centering
  \footnotesize
  \setlength{\tabcolsep}{6pt}
  \begin{tabular}{l c c}
    \toprule
    \textbf{Dataset} & \textbf{MLFI Storage} & \textbf{K-NN Storage} \\
    \midrule
    DomainNet      & 0.05 MB & 0.09 MB \\
    ImageNet-R     & 0.10 MB & 0.22 MB \\
    ImageNet-C     & 0.13 MB & 0.22 MB \\
    ImageNet-Mix   & 0.18 MB & 0.44 MB \\
    \bottomrule
  \end{tabular}
  \caption{Memory cost of MLFI and K-NN across different benchmarks.}
  \label{tab:knn_vs_mlfi_storage}
\end{table}

\subsection{Best Layer Searching and Memory Analysis}
\label{sec:best_layer_searching}
To identify the most informative layers for MLFI, we first evaluate the domain-prediction accuracy of each Transformer layer individually. Based on these scores, we perform an accuracy-guided greedy search: starting from the top-performing layers, we enumerate compact combinations and retain only those that further improve validation accuracy. This lightweight procedure effectively finds layer subsets that provide strong complementary cues while avoiding exhaustive search. The final selected layers are reported in ~\cref{tab:mlfi_layers}.

\minisection{Comparison with K-NN Storage.}
For reference, we also estimate the storage cost of the K-NN domain selector used in prior work.
In contrast to MLFI—which stores only a few averaged multi-layer features—K-NN must maintain several high-dimensional centroids in the full embedding space, leading to noticeably higher memory usage. The comparison is summarized in \cref{tab:knn_vs_mlfi_storage}.

\begin{table}[t]
  \centering
  \footnotesize
  \setlength{\tabcolsep}{10pt}
  \begin{tabular}{l c}
    \toprule
    \textbf{Dataset} & \textbf{Optimal $\lambda$} \\
    \midrule
    DomainNet      & 0.2 \\
    ImageNet-R     & 1.0 \\
    ImageNet-C     & 0.5 \\
    ImageNet-Mix   & 1.0 \\
    \bottomrule
  \end{tabular}
  \caption{Optimal alignment-loss weight $\lambda$ across benchmarks.}
  \label{tab:lambda_settings}
\end{table}

\subsection{Alignment Loss Hyperparameters}
\label{sec:lambda}

\Cref{tab:lambda_settings} presents the optimal alignment-loss weight $\lambda$ for each benchmark. The preferred values vary slightly across datasets, suggesting that a moderate level of alignment tends to work well in practice.

\begin{table*}[t]
  \centering
  \footnotesize
  \resizebox{0.9\textwidth}{!}{
  \begin{tabular}{l c c c c}
    \toprule
    \textbf{Method} & \textbf{Params (Trainable/Total, M)} & \textbf{GPU Mem (GB)} & \textbf{Batch Time (s)} & \makecell{\textbf{$A_T$} \textbf{($\Delta$)}} \\
    \midrule
    L2P~\cite{wang2022learning}           & 0.200 / 149.82 & 16.56 & 0.21 & 65.27 (+15.62) \\
    Dual-Prompt~\cite{wang2022dualprompt} & 1.098 / 150.72 & 18.39 & 0.61 & 71.10 (+9.79) \\
    CODA-Prompt~\cite{smith2023coda}      & 5.914 / 155.53 & 18.16 & 0.60 & 69.07 (+11.82) \\
    CPrompt~\cite{gao2024consistent}      & 5.917 / 155.54 & 40.27 & 1.34 & 74.02 (+6.87) \\
    \midrule
    S-Prompts~\cite{wang2022s}            & 0.307 / 149.93  & 14.10 & 0.38 & 41.23 (+39.66) \\
    PINA~\cite{wang2024non}               & 4.757 / 154.38  & 13.99 & 0.29 & 63.80 (+17.09) \\
    CP-Prompt~\cite{feng2024cp}           & 5.887 / 155.51  & 12.60 & 0.26 & 60.98 (+19.91) \\
    C-Prompts~\cite{liu2024compositional} & 10.559 / 160.18 &  9.79 & 0.32 & 74.57 (+6.32) \\
    KA-Prompts~\cite{xu2025componential}  & 10.559 / 160.18 & 19.54 & 0.60 & 75.12 (+5.77) \\
    \midrule
    LAVA-share (ours) & 4.749 / 154.37  & 16.45 & 0.46 & 80.27 (+0.62) \\
    LAVA (ours)       & 7.053 / 156.68  & 15.33 & 0.48 & 80.89 (Best) \\
    \bottomrule
  \end{tabular}%
  }
\caption{
Comparison of the number of parameters, overhead, and performance with the state-of-the-art on ImageNet-R.
``Params'' reports trainable/total parameters (in M) on top of a frozen backbone with 149.62M parameters;
``GPU Mem'' is the peak GPU memory in GB (MB$/1024$);
``Batch Time'' is the per-batch running time in seconds;
$A_T$ is the final average task accuracy, and $\Delta$ denotes the accuracy gap w.r.t.\ LAVA, 
computed as $\Delta = A_T(\text{LAVA}) - A_T(\text{method})$.
}
\label{tab:complexity_core_with_AT_last}
\end{table*}

\begin{table*}[t]
  \centering
  \footnotesize
  \resizebox{0.85\textwidth}{!}{
  \begin{tabular}{l c c c}
    \toprule
    \textbf{Method} &
    \textbf{Active Params@Inf. (Extra/Total, M, +\%)} &
    \textbf{GPU Mem (GB)} &
    \textbf{Latency (s)} \\
    \midrule
    L2P~\cite{wang2022learning}            & 0.200 / 149.82   (+0.13\%) & 2.00 & 0.04 \\
    Dual-Prompt~\cite{wang2022dualprompt}  & 1.098 / 150.72   (+0.73\%) & 4.04 & 0.36 \\
    CODA-Prompt~\cite{smith2023coda}       & 5.914 / 155.53   (+3.95\%) & 4.07 & 0.32 \\
    CPrompt~\cite{gao2024consistent}       & 2.688 / 152.31   (+1.80\%) & 2.23 & 0.10 \\
    \midrule
    S-Prompts~\cite{wang2022s}             & 0.020 / 149.64   (+0.01\%) & 3.68 & 0.90 \\
    PINA~\cite{wang2024non}                & 0.317 / 149.94   (+0.21\%) & 2.95 & 1.23 \\
    CP-Prompt~\cite{feng2024cp}            & 0.385 / 150.01   (+0.26\%) & 2.52 & 0.45 \\
    C-Prompts~\cite{liu2024compositional}  & 10.559 / 160.18  (+7.06\%) & 3.71 & 0.32 \\
    KA-Prompts~\cite{xu2025componential}   & 10.559 / 160.18  (+7.06\%) & 5.25 & 0.32 \\
    \midrule
    LAVA-share (ours)  & 0.317 / 149.94   (+0.21\%) & 2.44 & 0.37 \\
    LAVA (ours)        & 0.470 / 150.09   (+0.31\%) & 2.49 & 0.38 \\
    \bottomrule
  \end{tabular}
  }
  \caption{
  Inference efficiency comparison with the state-of-the-art on ImageNet-R.
  ``Active Params@Inf.'' reports, for each method, the extra/total active parameters
  (in M) on top of a frozen backbone with 149.62M parameters, and the percentage
  denotes the relative increase over the backbone;
  ``GPU Mem'' is the peak GPU memory usage in GB;
  ``Latency'' is the per-batch inference time in seconds.
  }
  \label{tab:inference_efficiency}
\end{table*}

\begin{algorithm*}[t]
    \caption{LAVA Training and Inference Algorithm}
    \label{alg:lava_framework}
    \textbf{Input}: Training sets $\mathcal{X}_{1},\cdots,\mathcal{X}_{T}$, test sets $\mathcal{Z}_{1},\cdots,\mathcal{Z}_{T}$, class names for label set $\mathcal{C}$. \\
    \textbf{Output}: The predicted labels of test images.
    
    \begin{algorithmic}[1]
    \STATE \textit{// Initialization Phase}
    \STATE Pre-compute static Text-based Anchors $\mathcal{A}^{\text{Text}}$ using the frozen text encoder $G$.
    \STATE Initialize learnable pools with random parameters:
Prompts $\mathcal{P}$, Visual Anchors $\mathcal{A}^{\text{Vis}}$,
Prototype Anchors $\mathcal{V}^{\text{proto}}$, and Classifiers $\Psi$.
    
    \STATE \textit{// Training Phase}
    \FOR {$t$ in $[1,T]$}
        \FOR {each $(x_i, y_i)$ in $\mathcal{X}_t$}
            \STATE Compute the prompt-adapted feature $\mathbf{g}_i$ using prompt $\mathbf{P}_{(t)}$, per \cref{eq:prompt}.
            \STATE Compute structural loss $\mathcal{L}_{\text{Struct}}$ via the VL-RSA module, using \cref{eq:r_v,eq:ryi,eq:rgi,eq:kl}.
            \STATE Compute aggregated feature $\mathbf{f}_i$ via the CA-CDFA module, per \cref{eq:global_fused,eq:global_alpha}.
            \STATE Optimize learnable parameters for domain $t$ ($\mathbf{P}_{(t)}, \dots, \psi^{(t)}$) by minimizing the combined loss $\mathcal{L}$ from \cref{eq:combine_loss}.
        \ENDFOR
        \STATE Using the MLFI module, compute the domain prototype $\boldsymbol{\mu}_{(t)}$ for domain $t$ on $\mathcal{X}_t$ per \cref{eq:concat,eq:prototype}.
    \ENDFOR

    \STATE \textit{// Inference Phase}
    \STATE Get the test set $\mathcal{Z}_{1 \sim T} = \bigcup_{\tau = 1}^{T} \mathcal{Z}_{\tau}$.
    \FOR {$z$ in $\mathcal{Z}_{1 \sim T}$}
        \STATE Compute multi-level feature $F(z)$ using the MLFI extractor per \cref{eq:concat}.
        \STATE Identify domain label $s = \underset{1 \le t \le T}{\arg\max} \text{cos}(F(z), \boldsymbol{\mu}_{t})$.
        \STATE Compute prompt-adapted feature $\mathbf{g}_{z}$ using components from domain $s$, per \cref{eq:prompt}.
        \STATE Compute aggregated feature $\mathbf{f}_{z}$ using components up to domain $s$, per \cref{eq:global_fused,eq:global_alpha}.
        \STATE Predict label $\hat{y} = \psi^{(s)}(\mathbf{f}_{z})$ using the classifier for domain $s$.
    \ENDFOR
    \end{algorithmic}
\end{algorithm*}

\subsection{Computational Overhead and Efficiency}
\label{sec:parameter_comparison}

To better understand the computational profile of LAVA, we provide a detailed comparison of training and inference overhead on ImageNet-R in \cref{tab:complexity_core_with_AT_last} and \cref{tab:inference_efficiency}. These results complement the main paper by quantifying how LAVA trades parameter cost and runtime for accuracy.

\minisection{Training-time parameter and memory cost.}
\Cref{tab:complexity_core_with_AT_last} reports the number of trainable parameters, peak GPU memory, and per-batch training time for each method, together with the final average task accuracy $A_T$.
On ImageNet-R, LAVA achieves the best performance with $A_T = 80.89\%$, while introducing only 7.05M trainable parameters on top of a frozen 149.62M backbone.
This represents a substantial reduction in trainable parameters compared to parameter-heavy methods such as C-Prompts and KA-Prompts, both of which require 10.56M trainable parameters but still lag behind LAVA by \textbf{+6.32\%} and \textbf{+5.77\%} in $A_T$, respectively.
Even the more compact \textit{LAVA-share} variant, with 4.75M trainable parameters, already outperforms all baselines, reaching $80.27\%$ and remaining only \textbf{0.62\%} behind the full LAVA model.
Overall, LAVA and \textit{LAVA-share} attain these accuracy gains while keeping peak GPU memory and per-batch training time competitive with or more efficient than existing prompt-based baselines.

\minisection{Inference-time efficiency.}
\Cref{tab:inference_efficiency} further evaluates the runtime footprint at inference.
Here, we report the number of active parameters, including both the extra parameters 
introduced by each method and the resulting total parameter count at test time,
as well as the peak GPU memory and per-batch latency.
Despite using 7.05M trainable parameters during training, LAVA activates only 0.47M extra parameters at inference, corresponding to a \textbf{0.31\%} increase over the backbone.
This overhead is an order of magnitude smaller than that of C-Prompts and KA-Prompts, which each require 10.56M extra active parameters (\textbf{+7.06\%}).
The \textit{LAVA-share} variant is even more lightweight, with just 0.32M extra active parameters (\textbf{+0.21\%}) while preserving most of the performance.
Consequently, both LAVA and \textit{LAVA-share} maintain a lightweight inference-time footprint, with GPU memory usage and latency that are on par with or better than other strong baselines, while still delivering a clear accuracy advantage.

\minisection{Overall parameter efficiency and accuracy--cost trade-off.}
A key aspect of LAVA’s design is its \emph{per-stage} efficiency: although the total number of trainable parameters grows with the number of domains, the parameters updated at each incremental stage remain modest compared to other high-performing, parameter-expanding methods.
Practically, this leads to improved training efficiency, since only a small subset of the overall parameters needs to be updated when a new domain arrives.
Overall, these results demonstrate that LAVA’s state-of-the-art performance is not achieved by merely increasing parameter counts or computation: LAVA uses a moderate training-time parameter budget together with an extremely small inference-time overhead, and converts these resources into substantial accuracy gains, achieving a superior balance between accuracy, memory footprint, and runtime cost across both training and inference.

\begin{figure*}[t]
    \centering
    \includegraphics[width=0.9\textwidth]{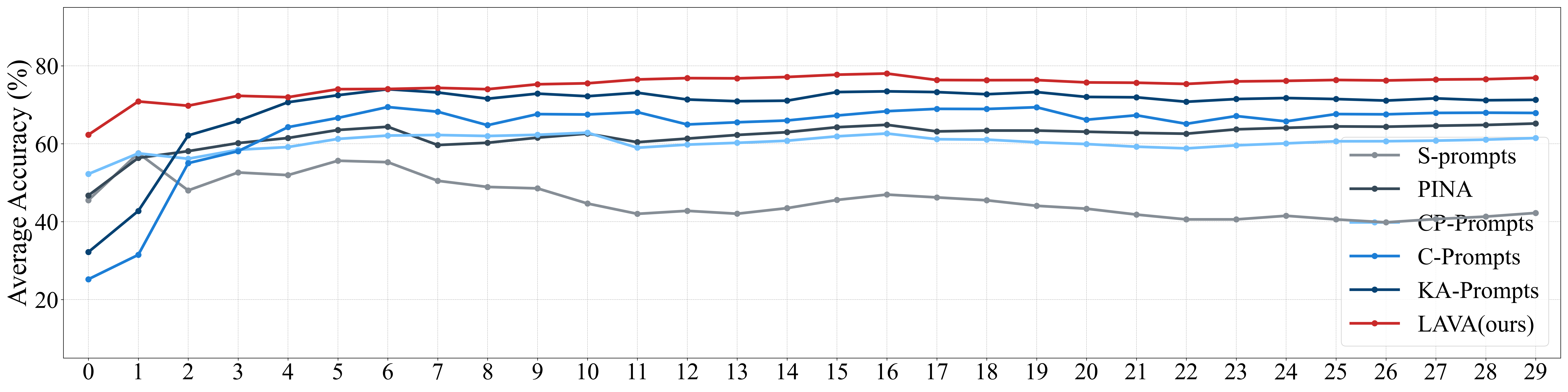}
    \caption{ImageNet-Mix average accuracy on seen domains.}
    \label{fig:imagenet_mix}
\end{figure*}

\subsection{Further Details of the Training and Inference}
\cref{alg:lava_framework} introduces how to train and test our Language-Anchored Visual Alignment (LAVA) framework. The symbols correspond to those used in the Problem Formulation subsection and the Method section.

\subsection{Visualization Results}
\begin{figure}[t]
    \centering
    \includegraphics[width=0.45\textwidth]{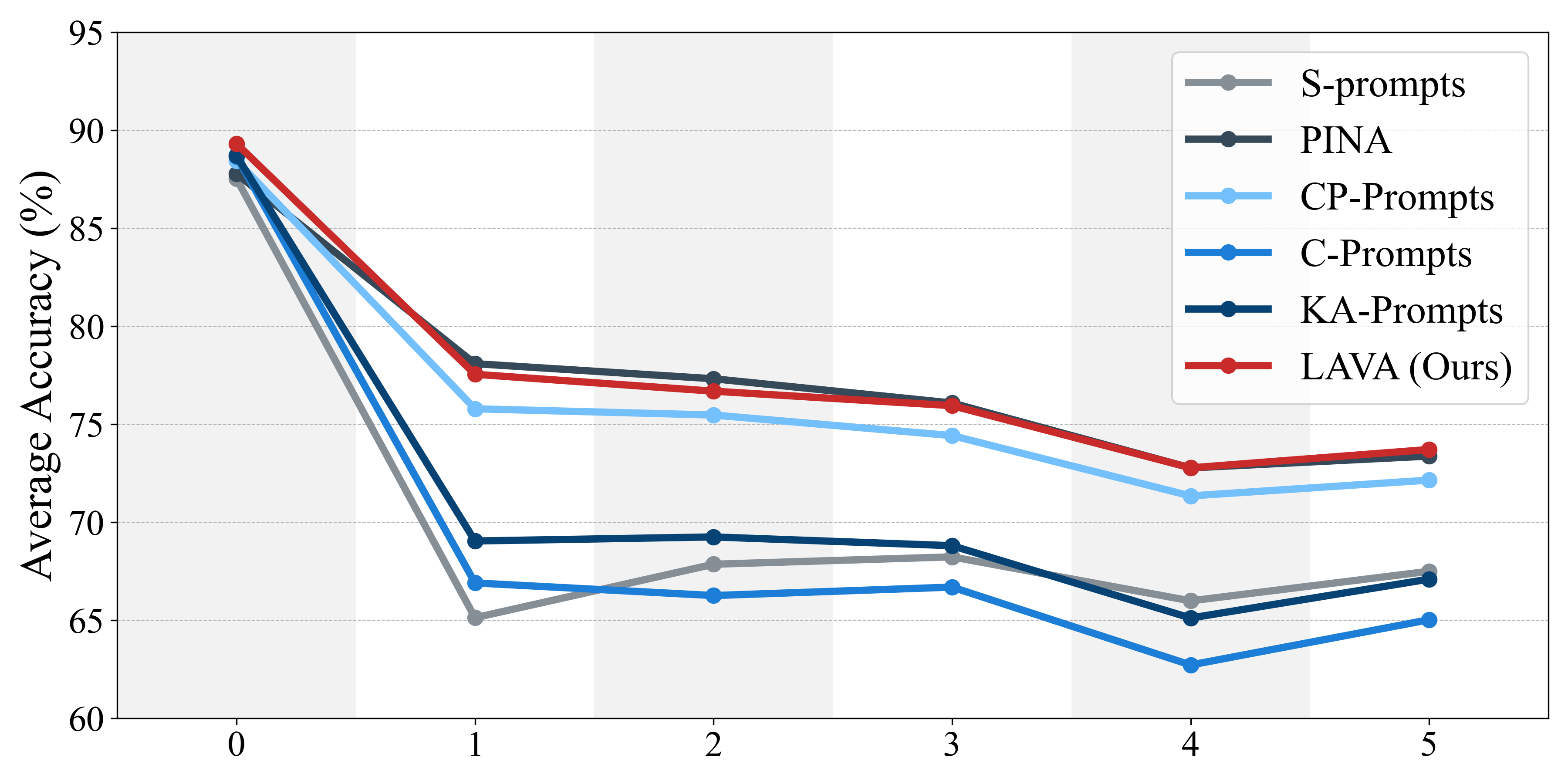}
    \caption{DomainNet average accuracy on seen domains.}
    \label{fig:domainnet}
\end{figure}

In addition to the main paper, we also provide the performance trends on seen domains for two additional benchmarks, i.e., DomainNet and ImageNet-Mix, as illustrated in \cref{fig:domainnet} and \cref{fig:imagenet_mix}.

As shown in \cref{fig:domainnet}, LAVA sets a new state-of-the-art on the DomainNet benchmark, outperforming the prior SOTA method, PINA. This result is particularly compelling as the experiment utilizes the \textit{Real}-first domain sequence, a setting known to be most favorable for PINA due to its architectural reliance on the base domain. Despite this, LAVA secures a \textbf{+0.46\%} advantage. The performance gap widens significantly in other permutations; for instance, LAVA's lead increases to a substantial \textbf{+4.93\%} when \textit{Quickdraw} is the initial domain.

LAVA's robustness is further demonstrated on the more demanding ImageNet-Mix benchmark (\cref{fig:imagenet_mix}), which comprises 30 sequential domain shifts. Here, LAVA establishes a decisive and consistent lead over all baseline methods throughout the entire learning process. This significant performance margin underscores LAVA's superior generalization and robustness in complex, long-sequence continual learning scenarios.

\subsection{Limitations and Future Directions}
\label{sec:limitations}

While LAVA performs strongly across diverse benchmarks, several limitations remain. 
First, the method implicitly assumes a certain degree of semantic structural consistency across domains, as the relative alignment mechanism relies on semantic relationships in the embedding space remaining reasonably stable.
In practice, this assumption generally holds for natural-image benchmarks; however, if the model is applied to domains with extreme discrepancies in visual style, statistical properties, or semantic ontology—such as highly stylized or artificially constructed domains with little correspondence to natural-image semantics—the shared language-anchored structure may become less reliable, potentially reducing adaptation effectiveness and degrading performance on those domains.

Second, although only a small number of parameters are updated at each stage, the cumulative parameter count still grows linearly with the number of domains. 
Reducing redundancy by merging pools for semantically similar domains or reusing pools when domain similarity is high represents a promising direction for improving long-horizon scalability.

Looking ahead, future work could pursue more robust domain-invariant structures and more scalable parameterization strategies that better leverage cross-domain commonality. 
Such developments may enhance the deployment of incremental models in dynamic environments and deepen our understanding of stability–plasticity trade-offs in large-scale vision–language systems.

%% file: main.bib
@String(ICLR = {Int. Conf. Learn. Represent.})

@String(AAAI = {AAAI})

@String(ICLR  = {ICLR})

@article{diko2024laguna,
  title={LAGUNA: LAnguage Guided UNsupervised Adaptation with structured spaces},
  author={Diko, Anxhelo and Furnari, Antonino and Cinque, Luigi and Farinella, Giovanni Maria},
  journal={arXiv preprint arXiv:2411.15557},
  year={2024}
}

@inproceedings{wang2024non,
  title={Non-exemplar domain incremental learning via cross-domain concept integration},
  author={Wang, Qiang and He, Yuhang and Dong, Songlin and Gao, Xinyuan and Wang, Shaokun and Gong, Yihong},
  booktitle={European Conference on Computer Vision},
  pages={144--162},
  year={2024},
  organization={Springer}
}

@inproceedings{feng2024cp,
  title={CP-Prompt: Composition-Based Cross-modal Prompting for Domain-Incremental Continual Learning},
  author={Feng, Yu and Tian, Zhen and Zhu, Yifan and Han, Zongfu and Luo, Haoran and Zhang, Guangwei and Song, Meina},
  booktitle={Proceedings of the 32nd ACM International Conference on Multimedia},
  pages={2729--2738},
  year={2024}
}

@article{xu2025componential,
  title={Componential Prompt-Knowledge Alignment for Domain Incremental Learning},
  author={Xu, Kunlun and Zou, Xu and Hua, Gang and Zhou, Jiahuan},
  journal={arXiv preprint arXiv:2505.04575},
  year={2025}
}

@article{cannistraci2023bricks,
  title={From bricks to bridges: Product of invariances to enhance latent space communication},
  author={Cannistraci, Irene and Moschella, Luca and Fumero, Marco and Maiorca, Valentino and Rodol{\`a}, Emanuele},
  journal={arXiv preprint arXiv:2310.01211},
  year={2023}
}

@inproceedings{diko2024semantically,
  title={Semantically guided representation learning for action anticipation},
  author={Diko, Anxhelo and Avola, Danilo and Prenkaj, Bardh and Fontana, Federico and Cinque, Luigi},
  booktitle={European Conference on Computer Vision},
  pages={448--466},
  year={2024},
  organization={Springer}
}

@article{wang2022s,
  title={S-prompts learning with pre-trained transformers: An occam’s razor for domain incremental learning},
  author={Wang, Yabin and Huang, Zhiwu and Hong, Xiaopeng},
  journal={Advances in Neural Information Processing Systems},
  volume={35},
  pages={5682--5695},
  year={2022}
}

@inproceedings{dunlap2023using,
  title={Using Language to Extend to Unseen Domains.},
  author={Dunlap, Lisa and Mohri, Clara and Guillory, Devin and Zhang, Han and Darrell, Trevor and Gonzalez, Joseph E and Raghunathan, Aditi and Rohrbach, Anna},
  year={2023},
  organization={International Conference on Learning Representations (ICLR)}
}

@inproceedings{gokhale2021attribute,
  title={Attribute-guided adversarial training for robustness to natural perturbations},
  author={Gokhale, Tejas and Anirudh, Rushil and Kailkhura, Bhavya and Thiagarajan, Jayaraman J and Baral, Chitta and Yang, Yezhou},
  booktitle={Proceedings of the AAAI Conference on Artificial Intelligence},
  volume={35},
  number={9},
  pages={7574--7582},
  year={2021}
}

@inproceedings{huang2023sentence,
  title={A sentence speaks a thousand images: Domain generalization through distilling clip with language guidance},
  author={Huang, Zeyi and Zhou, Andy and Ling, Zijian and Cai, Mu and Wang, Haohan and Lee, Yong Jae},
  booktitle={Proceedings of the IEEE/CVF International Conference on Computer Vision},
  pages={11685--11695},
  year={2023}
}

@inproceedings{min2022grounding,
  title={Grounding visual representations with texts for domain generalization},
  author={Min, Seonwoo and Park, Nokyung and Kim, Siwon and Park, Seunghyun and Kim, Jinkyu},
  booktitle={European conference on computer vision},
  pages={37--53},
  year={2022},
  organization={Springer}
}

@article{wang2024landa,
  title={Landa: Language-guided multi-source domain adaptation},
  author={Wang, Zhenbin and Zhang, Lei and Wang, Lituan and Zhu, Minjuan},
  journal={arXiv preprint arXiv:2401.14148},
  year={2024}
}

@article{moschella2022relative,
  title={Relative representations enable zero-shot latent space communication},
  author={Moschella, Luca and Maiorca, Valentino and Fumero, Marco and Norelli, Antonio and Locatello, Francesco and Rodol{\`a}, Emanuele},
  journal={arXiv preprint arXiv:2209.15430},
  year={2022}
}

@article{maiorca2023latent,
  title={Latent space translation via semantic alignment},
  author={Maiorca, Valentino and Moschella, Luca and Norelli, Antonio and Fumero, Marco and Locatello, Francesco and Rodol{\`a}, Emanuele},
  journal={Advances in Neural Information Processing Systems},
  volume={36},
  pages={55394--55414},
  year={2023}
}

@article{norelli2023asif,
  title={Asif: Coupled data turns unimodal models to multimodal without training},
  author={Norelli, Antonio and Fumero, Marco and Maiorca, Valentino and Moschella, Luca and Rodola, Emanuele and Locatello, Francesco},
  journal={Advances in Neural Information Processing Systems},
  volume={36},
  pages={15303--15319},
  year={2023}
}

@article{devillers2021does,
  title={Does language help generalization in vision models?},
  author={Devillers, Benjamin and Choksi, Bhavin and Bielawski, Romain and VanRullen, Rufin},
  journal={arXiv preprint arXiv:2104.08313},
  year={2021}
}

@inproceedings{radford2021learning,
  title={Learning transferable visual models from natural language supervision},
  author={Radford, Alec and Kim, Jong Wook and Hallacy, Chris and Ramesh, Aditya and Goh, Gabriel and Agarwal, Sandhini and Sastry, Girish and Askell, Amanda and Mishkin, Pamela and Clark, Jack and others},
  booktitle={International conference on machine learning},
  pages={8748--8763},
  year={2021},
  organization={PmLR}
}

@inproceedings{peng2019moment,
  title={Moment matching for multi-source domain adaptation},
  author={Peng, Xingchao and Bai, Qinxun and Xia, Xide and Huang, Zijun and Saenko, Kate and Wang, Bo},
  booktitle={Proceedings of the IEEE/CVF international conference on computer vision},
  pages={1406--1415},
  year={2019}
}

@inproceedings{hendrycks2021many,
  title={The many faces of robustness: A critical analysis of out-of-distribution generalization},
  author={Hendrycks, Dan and Basart, Steven and Mu, Norman and Kadavath, Saurav and Wang, Frank and Dorundo, Evan and Desai, Rahul and Zhu, Tyler and Parajuli, Samyak and Guo, Mike and others},
  booktitle={Proceedings of the IEEE/CVF International Conference on Computer Vision},
  pages={8340--8349},
  year={2021}
}

@article{hendrycks2018benchmarking,
  title={Benchmarking neural network robustness to common corruptions and surface variations},
  author={Hendrycks, Dan and Dietterich, Thomas G},
  journal={arXiv preprint arXiv:1807.01697},
  year={2018}
}

@inproceedings{lomonaco2017core50,
  title={Core50: a new dataset and benchmark for continuous object recognition},
  author={Lomonaco, Vincenzo and Maltoni, Davide},
  booktitle={Conference on robot learning},
  pages={17--26},
  year={2017},
  organization={PMLR}
}

@article{liu2024compositional,
  title={Compositional prompting for anti-forgetting in domain incremental learning},
  author={Liu, Zichen and Peng, Yuxin and Zhou, Jiahuan},
  journal={International Journal of Computer Vision},
  volume={132},
  number={12},
  pages={5783--5800},
  year={2024},
  publisher={Springer}
}

@article{li2017learning,
  title={Learning without forgetting},
  author={Li, Zhizhong and Hoiem, Derek},
  journal={IEEE Transactions on Pattern Analysis and Machine Intelligence},
  volume={40},
  number={12},
  pages={2935--2947},
  year={2017},
  publisher={IEEE}
}

@article{kirkpatrick2017overcoming,
  title={Overcoming catastrophic forgetting in neural networks},
  author={Kirkpatrick, James and Pascanu, Razvan and Rabinowitz, Neil and Veness, Joel and Desjardins, Guillaume and Rusu, Andrei A and Milan, Kieran and Quan, John and Ramalho, Tiago and Grabska-Barwinska, Agnieszka and others},
  journal={Proceedings of the national academy of sciences},
  volume={114},
  number={13},
  pages={3521--3526},
  year={2017},
  publisher={National Acad Sciences}
}

@inproceedings{wang2022learning,
  title={Learning to prompt for continual learning},
  author={Wang, Zifeng and Zhang, Zizhao and Lee, Chen-Yu and Zhang, Han and Sun, Ruoxi and Ren, Xiaoqi and Su, Guolong and Perot, Vincent and Dy, Jennifer and Pfister, Tomas},
  booktitle={Proceedings of the IEEE/CVF Conference on Computer Vision and Pattern Recognition},
  pages={139--149},
  year={2022}
}

@inproceedings{wang2022dualprompt,
  title={Dualprompt: Complementary prompting for rehearsal-free continual learning},
  author={Wang, Zifeng and Zhang, Zizhao and Ebrahimi, Sayna and Sun, Ruoxi and Zhang, Han and Lee, Chen-Yu and Ren, Xiaoqi and Su, Guolong and Perot, Vincent and Dy, Jennifer and others},
  booktitle={European Conference on Computer Vision},
  pages={631--648},
  year={2022},
  organization={Springer}
}

@inproceedings{smith2023coda,
  title={Coda-prompt: Continual decomposed attention-based prompting for rehearsal-free continual learning},
  author={Smith, James Seale and Karlinsky, Leonid and Gutta, Vyshnavi and Cascante-Bonilla, Paola and Kim, Donghyun and Arbelle, Assaf and Panda, Rameswar and Feris, Rogerio and Kira, Zsolt},
  booktitle={Proceedings of the IEEE/CVF Conference on Computer Vision and Pattern Recognition},
  pages={11909--11919},
  year={2023}
}

@inproceedings{gao2024consistent,
  title={Consistent Prompting for Rehearsal-Free Continual Learning},
  author={Gao, Zhanxin and Cen, Jun and Chang, Xiaobin},
  booktitle={Proceedings of the IEEE/CVF Conference on Computer Vision and Pattern Recognition},
  pages={28463--28473},
  year={2024}
}

@inproceedings{liang2020we,
  title={Do we really need to access the source data? source hypothesis transfer for unsupervised domain adaptation},
  author={Liang, Jian and Hu, Dapeng and Feng, Jiashi},
  booktitle={International conference on machine learning},
  pages={6028--6039},
  year={2020},
  organization={PMLR}
}

@article{yang2021exploiting,
  title={Exploiting the intrinsic neighborhood structure for source-free domain adaptation},
  author={Yang, Shiqi and Van de Weijer, Joost and Herranz, Luis and Jui, Shangling and others},
  journal={Advances in neural information processing systems},
  volume={34},
  pages={29393--29405},
  year={2021}
}

@article{van2022three,
  title={Three types of incremental learning},
  author={Van de Ven, Gido M and Tuytelaars, Tinne and Tolias, Andreas S},
  journal={Nature Machine Intelligence},
  volume={4},
  number={12},
  pages={1185--1197},
  year={2022},
  publisher={Nature Publishing Group UK London}
}

@article{yoon2017lifelong,
  title={Lifelong learning with dynamically expandable networks},
  author={Yoon, Jaehong and Yang, Eunho and Lee, Jeongtae and Hwang, Sung Ju},
  journal={arXiv preprint arXiv:1708.01547},
  year={2017}
}

@article{hung2019compacting,
  title={Compacting, picking and growing for unforgetting continual learning},
  author={Hung, Ching-Yi and Tu, Cheng-Hao and Wu, Cheng-En and Chen, Chien-Hung and Chan, Yi-Ming and Chen, Chu-Song},
  journal={Advances in neural information processing systems},
  volume={32},
  year={2019}
}

@article{shi2023unified,
  title={A unified approach to domain incremental learning with memory: Theory and algorithm},
  author={Shi, Haizhou and Wang, Hao},
  journal={Advances in Neural Information Processing Systems},
  volume={36},
  pages={15027--15059},
  year={2023}
}

@article{jeeveswaran2024gradual,
  title={Gradual divergence for seamless adaptation: A novel domain incremental learning method},
  author={Jeeveswaran, Kishaan and Arani, Elahe and Zonooz, Bahram},
  journal={arXiv preprint arXiv:2406.16231},
  year={2024}
}

@article{kumar2022fine,
  title={Fine-tuning can distort pretrained features and underperform out-of-distribution},
  author={Kumar, Ananya and Raghunathan, Aditi and Jones, Robbie and Ma, Tengyu and Liang, Percy},
  journal={arXiv preprint arXiv:2202.10054},
  year={2022}
}

@inproceedings{wortsman2022robust,
  title={Robust fine-tuning of zero-shot models},
  author={Wortsman, Mitchell and Ilharco, Gabriel and Kim, Jong Wook and Li, Mike and Kornblith, Simon and Roelofs, Rebecca and Lopes, Raphael Gontijo and Hajishirzi, Hannaneh and Farhadi, Ali and Namkoong, Hongseok and others},
  booktitle={Proceedings of the IEEE/CVF conference on computer vision and pattern recognition},
  pages={7959--7971},
  year={2022}
}
